\documentclass[11pt]{article}

\usepackage[final]{acl}

\usepackage{times}
\usepackage{latexsym}

\usepackage[T1]{fontenc}
\usepackage[utf8]{inputenc}

\usepackage{microtype}

\usepackage{inconsolata}

\usepackage{graphicx}
\usepackage{fontawesome}

\usepackage{amsmath, amsfonts, amssymb} %lan
\usepackage{dsfont}
\usepackage{bbm}
\usepackage{booktabs}
\usepackage{algorithm}
\usepackage{algpseudocode}
\usepackage{multirow}
\usepackage{pifont}
\usepackage{enumitem}

\usepackage[table,xcdraw]{xcolor}
\newcommand{\tcr}{\textcolor{red}}
\definecolor{myDarkYellow}{rgb}{0.8, 0.6, 0.1}

\definecolor{myDarkGreen}{rgb}{0.1, 0.5, 0.2}

\definecolor{myDarkPurple}{rgb}{0.3, 0.1, 0.5}

\title{CSPO: Alleviating Reward Ambiguity for Structured Table-to-LaTeX Generation}

\author{
Yunfan Yang\textsuperscript{1}\thanks{~~Work done during internship at Microsoft Research Asia.},
Cuiling Lan\textsuperscript{2}\thanks{~~Corresponding authors.},
Jitao Sang\textsuperscript{1,3}\footnotemark[2], 
Yan Lu\textsuperscript{2}\\
\textsuperscript{1}Beijing Key Laboratory of Traffic Data Mining and Embodied Intelligence, Beijing Jiaotong University,\\
\textsuperscript{2}Microsoft Research Asia,
\textsuperscript{3}Peng Cheng Laboratory\\
\{yunfanyang, jtsang\}@bjtu.edu.cn,
\{culan, yanlu\}@microsoft.com\\
}

\newcommand{\ourss}{CSPO~} %MRACO~}%AutoTable~}
\newcommand{\ours}{CSPO} %MRACO}

\newcommand{\benchmarkno}{TableTex} %no space
\newcommand{\ieno}{\textit{i}.\textit{e}.}
\newcommand{\egno}{\textit{e}.\textit{g}.} %there is no space
\setlist[itemize]{leftmargin=0.6em,topsep=0.0em,itemsep=0.0em,parsep=0.0em}

\begin{document}
\maketitle
\begin{abstract}
Tables contain rich structured information, yet when stored as images their contents remain ``locked'' within pixels. Converting table images into LaTeX code enables faithful digitization and reuse, but current multimodal large language models (MLLMs) often fail to preserve structural, style, or content fidelity. 
Conventional post-training with reinforcement learning (RL) typically relies on a single aggregated reward, leading to reward ambiguity that conflates multiple behavioral aspects and hinders effective optimization. 
We propose \textbf{C}omponent-\textbf{S}pecific \textbf{P}olicy \textbf{O}ptimization (\ours), an RL framework that disentangles optimization across LaTeX tables components—structure, style, and content. In particular, \ourss assigns component-specific rewards and backpropagates each signal only through the tokens relevant to its component, alleviating reward ambiguity and enabling targeted component-wise optimization. To comprehensively assess performance, we introduce a set of hierarchical evaluation metrics. 
Extensive experiments demonstrate the effectiveness of \ours, underscoring the importance of component-specific optimization for reliable structured generation. Our code is available at \url{https://github.com/microsoft/CSPO}.
\end{abstract}
%Extensive experiments validate the effectiveness of our design, and \textcolor{red}{our 7B model significantly outperforms GPT-4o and Qwen2.5-VL-7B/72B}, setting a new state of the art in table image-to-LaTeX conversion.
%Extensive experiments validate the effectiveness of our design, and \textcolor{red}{our 7B model significantly outperforms GPT-4o and Qwen2.5-VL-7B/72B}, setting a new state of the art in table image-to-LaTeX conversion.

\section{Introduction}

\begin{figure}[t]
    \centering
    %\vspace{-3mm}
    \includegraphics[width=0.98\linewidth]{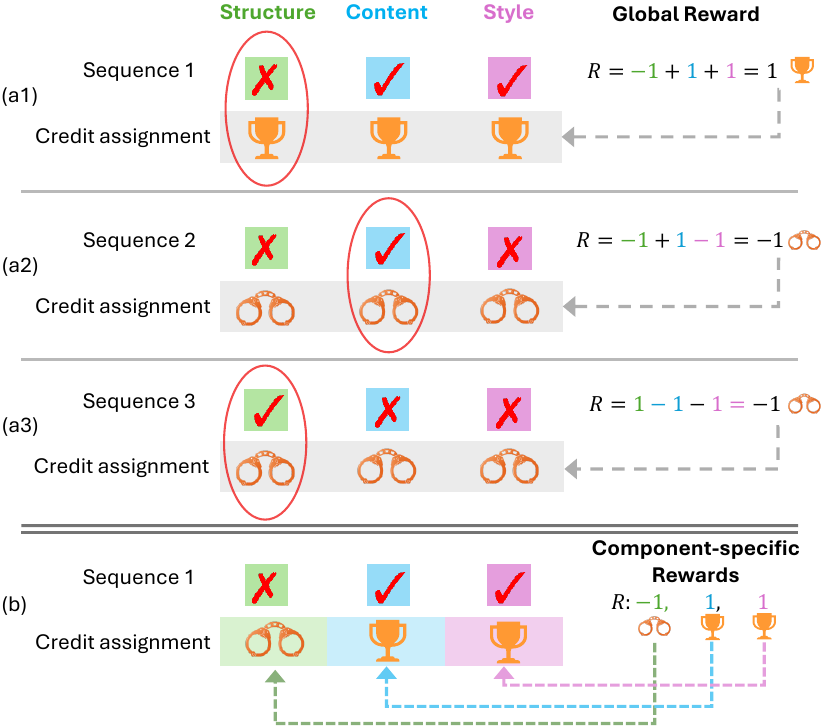}
    %\vspace{-5mm}
    \caption{Motivating example of reward ambiguity in table image-to-LaTeX generation. \protect\includegraphics[height=1em]{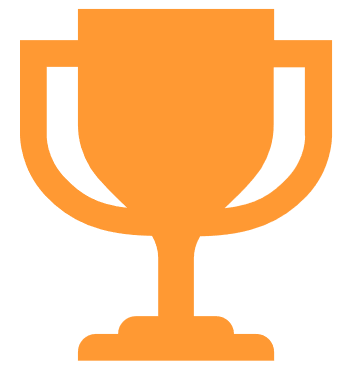} denotes a positive reward, while \protect\includegraphics[height=0.8em]{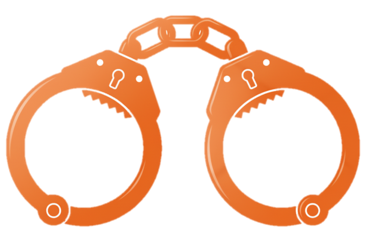} denotes a penalty. (a) Given a table image, multiple LaTeX sequences are generated with varied errors in structure, content, and style (errors marked with \textcolor{red}{\ding{55}}). Using a single global reward leads to \textbf{reward ambiguity}, where (a1) an incorrect \emph{structure} is erroneously reinforced, (a2) correct \emph{content} is unfairly penalized, and (a2) and (a3) receive identical rewards despite differing component fidelity. (b) Our method mitigates this by decomposing the LaTeX code into functional components and \textbf{assigning component-specific rewards for targeted optimization, thereby alleviating reward ambiguity}.}
\label{fig:motivation}
\end{figure}
% \begin{figure}[t]
%     \centering
%     %\vspace{-3mm}
%     \includegraphics[width=0.98\linewidth]{figures/Motivation.pdf}
%     %\vspace{-5mm}
%     \caption{Motivating example of reward ambiguity in table image-to-LaTeX generation. (a) Given a table image, a few LaTeX sequences are generated with varying errors (marked by \textcolor{red}{\ding{55}}) in structure, content, and style. Using a global aggregated reward leads to \textbf{reward ambiguity}, where (a1) an incorrect \emph{structure} is mistakenly reinforced, (a2) correct \emph{content} is wrongly penalized, and (a2) and (a3) receive identical rewards despite differing component fidelity. (b) Our method: decompose LaTeX code into functional components and \textbf{assign dedicated rewards to each for targeted optimization, alleviating reward ambiguity}.}
% \label{fig:motivation}
% \end{figure}
% \textcolor{red}{\ding{51}} 

\begin{figure*}[t]
    \centering
    \includegraphics[width=0.98\linewidth]{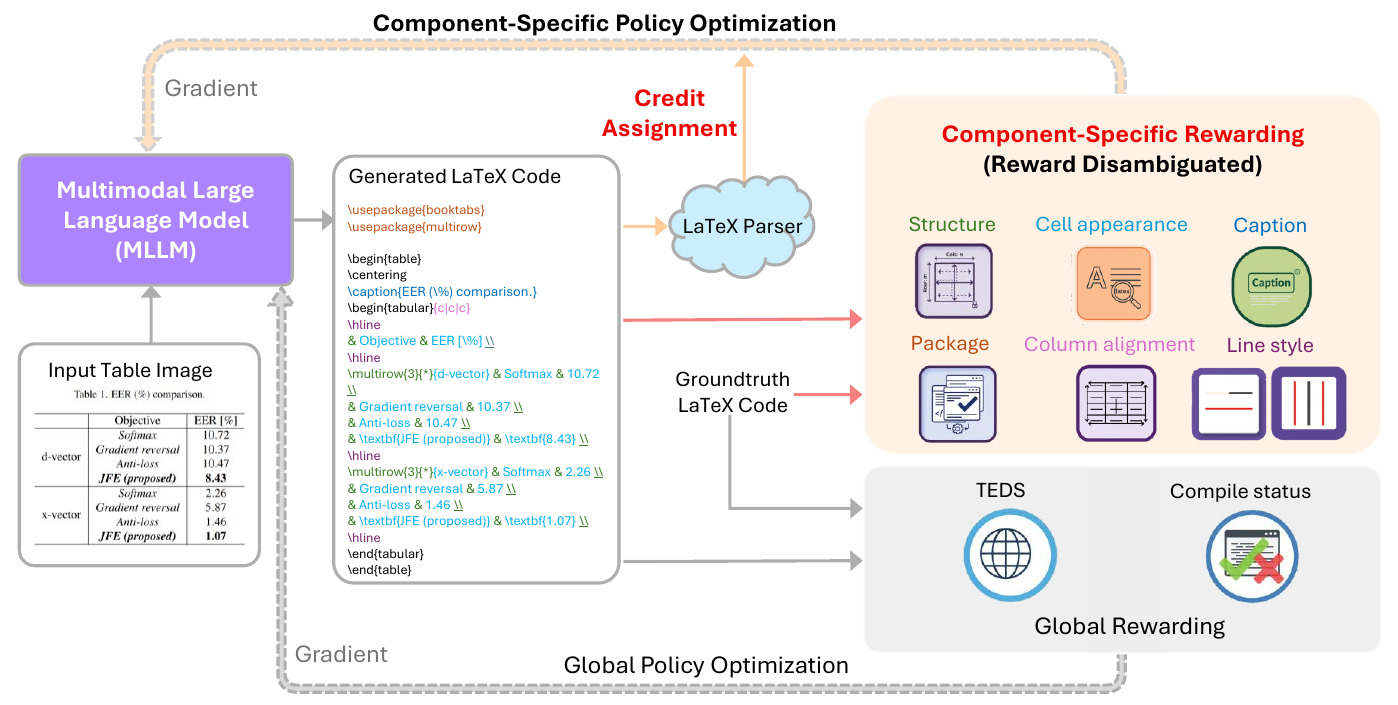} %Framework.pdf
    %\vspace{-5mm}
        \caption{
Overview of \textbf{Component-Specific Policy Optimization (CSPO)}. Particularly, CSPO decomposes each generated code sequence into functional components (\egno, structure, cell appearance, caption, package inclusion, alignment, and line style) using a LaTeX parser. It conducts \textbf{component-specific rewarding} by assessing each component's fidelity (with a strong LLM as the judge), performs \textbf{component-specific credit assignment} and \textbf{optimization},  effectively mitigating reward ambiguity in table image-to-LaTeX generation.
%\textbf{balances component rewards adaptively} (see Section~\ref{sec:approach}), for component-specific policy optimization,  effectively mitigating reward ambiguity in table image-to-LaTeX generation.
%Each component receives a dedicated reward for its own fidelity, alongside \textbf{reliable component-level credit assignment}, adaptive reward balancing for \textbf{component-specific policy optimization}, effectively mitigating reward ambiguity in table-to-LaTeX generation.  
%In addition, a global reward is used to capture overall sequence quality.
} %powered by component-specific rewarding, credict assignment, and dynamic reweighting
%\footnotetext{We denote different components with different colors in the code for illustration.}
\label{fig:framework}
\end{figure*}
%Overview of \textbf{Component-Specific Policy Optimization (CSPO)}. CSPO decomposes each generated code sequence into functional components (\egno, structure, cell appearance, caption, package inclusion, alignment, and line style) using a LaTeX parser. Each component receives a dedicated reward for its own fidelity, alongside \textbf{reliable component-level credit assignment}, adaptive reward balancing for \textbf{component-specific policy optimization}, effectively mitigating reward ambiguity in table-to-LaTeX generation. In addition, a global reward is used to capture overall sequence quality.
%Each component receives a dedicated reward for its own fidelity, enabling \textbf{reliable credit assignment} and \textbf{component-specific policy optimization} that effectively mitigates reward ambiguity in table-to-LaTeX generation. 

Scientific documents often contain complex tables that encapsulate critical data and insights \citep{gemelli2024datasets,xia2024docgenome,jiang2025latte}. However, when these tables are embedded in images—such as screenshots or PDFs—their structured information becomes locked within pixels, hindering data extraction, analysis, and reuse. Accurately converting table images into structured LaTeX code is therefore a crucial step toward  reliable digitalization and seamless editing.

Recent studies have explored the vision-to-code problem through specialized systems tailored for table understanding. LATTE \citep{jiang2025latte} introduces iterative error localization and correction, while DocGenome \citep{xia2024docgenome} fine-tunes Pix2Struct \citep{lee2023pix2struct} for table parsing. Beyond these task-specific efforts, multimodal large language models (MLLMs) such as GPT-4o and Qwen2.5-VL have demonstrated strong visual-to-text generalization, enabling zero-shot LaTeX generation. Nevertheless, both specialized systems and general MLLMs often introduce structural inconsistencies (\egno, incorrect cell merges), lose fine-grained formatting details (\egno, mismatched lines), content mistakes, or generate non-compilable code. This motivates the question: how can we effectively align MLLMs for this highly structured table generation?

Reinforcement learning (RL) has become a dominant paradigm for post-training alignment, yielding substantial improvements in reasoning, programming, and mathematical problem solving~\citep{ouyang2022training,qu2025survey,yu2025aligning,guo2025deepseek}. However, its use in table image-to-LaTeX generation remains largely underexplored. Unlike free-form language tasks (\egno, visual question answering), table-to-LaTeX generation presents distinct challenges, including multi-faceted fidelity (covering structure, content, style), hierarchical syntax (\egno, properly nested tabular and multicolumn structures), and execution sensitivity (the code part in Figure~\ref{fig:framework} shows an example of LaTeX code sequence).
%(compilation failure from minor syntax errors).

Existing RL approaches typically compute a single aggregated reward across the entire output sequence \citep{shao2024deepseekmath,yu2025dapo,ling2025table2latex}. For such a highly structured sequence generation task, this global aggregation is problematic, as it introduces \textbf{reward ambiguity—where fundamentally heterogeneous aspects such as structure, content, and style are collapsed into a single undifferentiated signal/reward.} Consequently, the model struggles to assign credit accurately, \textbf{leading to unreliable gradients and limited fidelity improvements}.
Figure~\ref{fig:motivation} illustrates this issue: in (a1), an incorrect structure component receives a positive global reward, mistakenly reinforcing the error; in (a2), the correct content component is wrongly penalized; (a2) and (a3) receive identical aggregated rewards despite differing in component fidelity, failing to distinguish good from bad performance in individual components.
%These examples underscore the need for component-specific reward attribution to enable fine-grained, targeted optimization in LaTeX generation.
%. It introduces \textbf{reward ambiguity, where the performance of fundamentally heterogeneous components— structure, content, style—is jointly assessed, resulting in a single, collapsed global reward signal.} As a result, the model cannot effectively distinguish errors arising from different components, leading to noisy gradient updates and limited fidelity improvements. 

To address this challenge, we propose Component-Specific Policy Optimization (CSPO), an RL framework specifically designed to mitigate reward ambiguity by assigning dedicated rewards to distinct functional components of LaTeX tables. As illustrated in Figure~\ref{fig:framework}, CSPO combines a global reward that captures overall output quality with component-specific rewards that disentangle structure, content, and style. During RL training, CSPO employs a LaTeX parser to decompose each generated code sequence into fine-grained functional components, including package imports, structure, cell appearance, captions, alignment, and line style. 
It then performs \textbf{component-specific rewarding} by evaluating each component's fidelity, conducts \textbf{component-specific credit assignment and optimization}, leading to more reliable component-level policy optimization. 
%It then performs \textbf{component-specific rewarding} by evaluating each component's fidelity, conducts \textbf{component-specific credit assignment}, and \textbf{adaptively balances component contributions} for more reliable component-level policy optimization. 
This approach ensures that improvements in one component are not overshadowed by errors in others, thereby enhancing the model's ability to generate faithful LaTeX code.
%Each component receives a component-specific reward, enabling targeted optimization. This ensures faithful credit assignment, preventing errors in one component from obscuring improvements in others, and vice versa, thereby enhancing the model's ability to generate faithful LaTeX code.

Furthermore, to enable more diagnostic and interpretable evaluation, we introduce a suite of \textbf{hierarchical metrics}. Beyond global similarity and compilation checks, these metrics separately measure structural correctness (\egno, row/column spans), content fidelity (\egno, cell values), and stylistic consistency (\egno, line style, font styles), providing granular feedback on model performance.
%at multiple granularity levels. 

%evaluating structural correctness (\egno, cell spanning), content fidelity, stylistic fidelity (\egno, font or line style). 

%\setlist[itemize]{leftmargin=0.6em,topsep=0.0em,itemsep=0.0em,parsep=0.0em} %final
Our contributions are summarized as follows:
\begin{itemize}
    \item  We identify \textbf{reward ambiguity} as a fundamental challenge in RL-based structured sequence generation for table-to-LaTeX conversion.
    \item We propose \ours, an effective RL framework that addresses reward ambiguity through \textbf{component-specific rewarding} and \textbf{explicit credit assignment}, enabling reliable and controllable \textbf{component-specific optimization}.
    %    \item We propose \ours, an effective RL framework that mitigates reward ambiguity through component-specific rewarding \textcolor{red}{and credit assignment, enabling reliable per-component optimization while maintaining coordinated behavior across components.}
    %\item We propose \ours, an effective RL framework that mitigates reward ambiguity through component-specific rewarding, credit assignment, and adaptive reward balancing, enabling reliable, per-component optimization with coordinated contributions.
    \item We introduce \textbf{hierarchical evaluation metrics} for comprehensive and diagnostic assessment, providing useful signals for guiding future table-to-LaTeX model design and optimization.
\end{itemize}
Experimental results demonstrate the effectiveness of our CSPO, highlighting the importance of addressing reward ambiguity and offering a general blueprint for structured sequence generation tasks.
%underscoring the importance of addressing reward ambiguity and offering a potential blueprint for future structured sequence generation tasks.
%. Our model surpasses both closed- and open-source baselines, establishing a new state of the art. 
%\ours~focuses on component-specific optimization for table image to LaTeX generation, offering a potential blueprint for future structured sequence generation tasks.
%\ours~ focuses on component-specific optimization, which may provide insights for other structured sequence generation tasks in future.
%Our experiments demonstrate that the models trained with \ourss achieve substantial gains in both structural, style, and content fidelity. Specifically, our 7B model, \ours-7B, surpasses closed-source systems (GPT-4o, Gemini-2.5 Flash) and open-source baselines (Qwen2.5-VL-7B/72B) by significant margins, setting a new state of the art for table image-to-LaTeX conversion.

\section{Related Work}
\label{sec:related}

\noindent\textbf{Table Image to Structured Markup.} Research on image-based table recognition has evolved from early detection and structure-parsing pipelines to end-to-end systems that directly map images to structured markup. Benchmarks such as PubTabNet \citep{zhong2020image} were pivotal in this transition, introducing large-scale supervision for image-to-HTML conversion. Encoder–decoder architectures (\egno, EDD~\citealp{zhong2020image}) focused on HTML/XML outputs, motivating subsequent methods in image-to-structure generation.
% and the Tree Edit Distance Similarity (TEDS) metric for structural fidelity evaluation

Some recent studies adopted LaTeX as the target format for its precise layout control, publication-quality rendering, and seamless integration with scientific workflows—capabilities that HTML lacks. DocGenome \citep{xia2024docgenome} fine-tuned Pix2Struct \citep{lee2023pix2struct} for table image-to-LaTeX conversion. LATTE \citep{jiang2025latte} introduced iterative refinement with delta-view correction to improve renderable LaTeX extraction from PDFs. Concurrent with our work, \citet{ling2025table2latex} post-train MLLMs with RL, using a single aggregated global reward signal.
Despite these advances, these models still struggle to faithfully reconstruct tables, often producing misaligned structures, inconsistent formatting, or incorrect cell content. 
In this work, we identify reward ambiguity as a key challenge limiting the effectiveness of RL-based post-training and propose CSPO to address it.
%Moreover, datasets such as that from DocGenome \cite{xia2024docgenome}, LATTE \cite{jiang2025latte}, and \cite{ling2025table2latex} remain limited: they exclude table captions and sample-wise compilation package specifications—both essential for producing context-complete and compilable LaTeX outputs. We aim to develop models that generate faithful, ready-to-use LaTeX that preserves structure, style, and content.

\noindent\textbf{Post-training Alignment and RL.} LLMs and MLLMs have advanced rapidly, showing strong generalization across diverse tasks.  
To further enhance domain-specific skills, RL-based post-training has become a widely adopted strategy for alignment and performance improvement~\citep{ouyang2022training,qu2025survey,yu2025aligning,guo2025deepseek,perera2025beyond,lai2025med,tang2025codereasoner,jia2025chartreasoner}. 
Methods such as GRPO \citep{shao2024deepseekmath} have shown effectiveness in mathematical reasoning~\citep{yu2025dapo}, program synthesis~\citep{tang2025codereasoner}, and multimodal analysis~\citep{zhou2025reinforced,lai2025med}. 

However, most approaches rely on a single aggregated reward that holistically evaluates outputs ~\citep{shao2024deepseekmath,ling2025table2latex}, leading to reward ambiguity and unreliable optimization when applied to table image-to-LaTeX generation. 
We address this limitation through disambiguated credit assignment for component-specific policy optimization.

\begin{figure*}[t]
    \centering
  \includegraphics[width=1\linewidth]{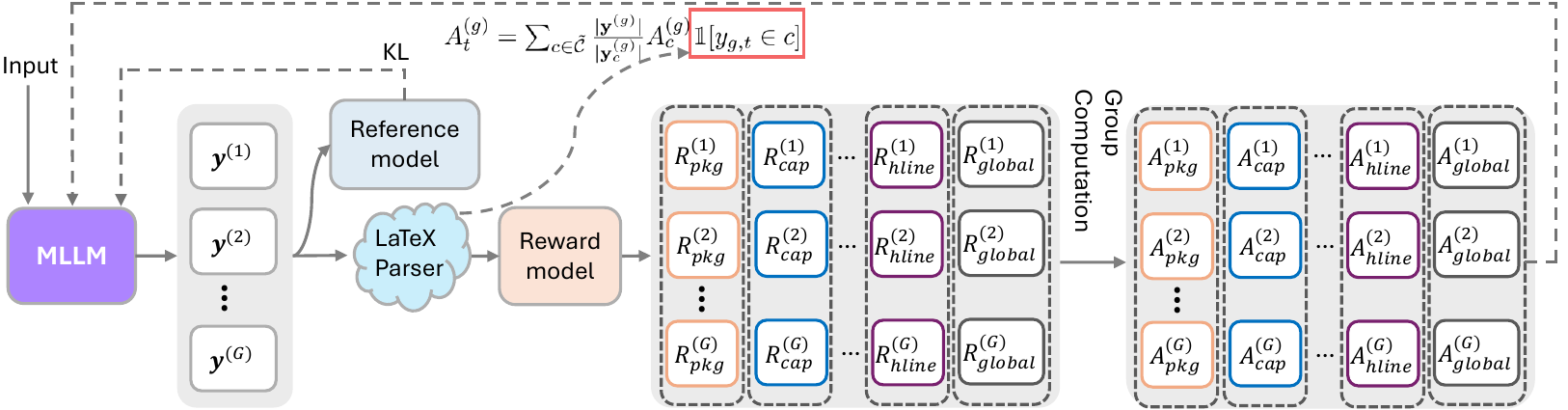} %CSPO.pdf}
    \vspace{-1em}
    \caption{Illustration of proposed CSPO algorithm. Each component-specific advantage $A_c^{(g)}$ ($c \in \tilde{\mathcal{C}}$) is calculated, and credited exclusively to its relevant tokens in rollout $g$. }
    %Component weights $w_c$ are dynamically updated based on reward signals for adaptive reward balancing.
    \label{fig:CSPO}
    \vspace{-0.5em}
\end{figure*}
%CSPO computes component-specific advantages $A_c^{(g)}$ ($c \in \tilde{\mathcal{C}}$)  

\section{Problem Formulation}

We study the task of \textbf{table image-to-LaTeX generation}, which aims to generate a compilable LaTeX code that faithfully reconstructs a given table image in terms of structure, content, and style. Given an input table image $\mathbf{x}$ and a LaTeX sequence $\mathbf{y} = (y_1, \ldots, y_T)$ of length $T$, a policy model $\pi_\theta$ defines a conditional distribution:
\begin{equation}
    \pi_\theta(\mathbf{y}|\mathbf{x}) = \prod_{t=1}^{T} \pi_\theta(y_t \mid y_{<t}, \mathbf{x}).
\end{equation}

To optimize generation quality, a natural approach is post-training the policy model with RL, which maximizes the expected reward:
\begin{equation}
    \mathcal{J}(\theta) = \mathbb{E}_{\mathbf{y}\sim\pi_\theta(\cdot|\mathbf{x})} \big[ R(\mathbf{y}, \mathbf{y}^*) \big],
\end{equation}
where $R(\mathbf{y}, \mathbf{y}^*)$ measures the consistency to the reference $\mathbf{y}^*$, \egno, via Tree‑Edit‑Distance‑based Similarity (TEDS) metric (see Appendix \ref{app:A}).
%Tree‑Edit‑Distance‑based Similarity (TEDS) measures overall quality by representing LaTeX programs as trees and computing a normalized tree-edit distance (see Appendix~A).

\section{Component-Specific Policy Optimization}
\label{sec:approach}

\noindent\textbf{Reward Ambiguity.} However, such a single holistic reward $R$ (\egno, TEDS) conflates heterogeneous aspects of model behavior—structure, content, and style— making it difficult to discern which parts of the output are correct or erroneous (see Figure~\ref{fig:motivation}). This reward ambiguity motivates our component-specific approach introduced next.

We propose Component-Specific Policy Optimization (CSPO), with the overall pipeline shown in Figure~\ref{fig:framework}. In particular, \ourss performs component-specific policy optimization through component decomposition, component-specific rewarding, credit assignment and optimization, enabling reliable credit attribution and policy updates for faithful LaTeX code generation.
%In particular, \ourss performs component-specific policy optimization through component decomposition, component-specific rewarding, credit assignment, and adaptive reward balancing, enabling reliable credit attribution and policy updates for faithful LaTeX code generation.
%It conducts component-specific policy optimization through component-specific rewarding, credit assignment, and adaptive reward balancing. This enables precise credit assignment and reliable policy updates. 
% It then assigns a dedicated reward to each component, enabling the model to distinguish which aspects are correct or require improvement. This fine-grained decomposition enables precise credit assignment  and component-specific policy updates. 
%CSPO further introduces adaptive reward balancing, dynamically reweighting components so that under-optimized ones gain emphasis while saturated ones are down-weighted, ensuring balanced optimization across all aspects.
%, an RL framework that moves beyond a single aggregated reward by disentangling structure, content, and style

\subsection{Component Decomposition}

LaTeX tables exhibit multi-dimensional fidelity, where correctness depends jointly on the \emph{structural layout} (\egno, rows, columns, merged cells), \emph{contents} (\egno, cell text and numbers),  \emph{stylistic attributes} (\egno, alignment, line styles, boldface), and \emph{compilability}. 

To facilitate disambiguated rewarding and credit assignment, we build a rule-based LaTeX parser to decompose each LaTeX sequence into seven functional components:
\begin{equation}
\begin{split}
\mathcal{C} = \left\{\text{pkg}, \text{cap}, \text{struct}, \text{cell-app}, \text{align}, \text{vline}, \text{hline}\right\},
\end{split}
\label{eq:C}
\end{equation}
which includes package dependencies (pkg), caption correctness (cap), structural organization (struct), cell appearance (cell-app), column alignment (align), and rule placement (vline, hline). Please see Appendix \ref{app:B} for more details. We illustrate the decomposition by marking different components with different colours over the generated LaTeX code in Figure~\ref{fig:framework}.  
%The code part in Figure~\ref{fig:framework} illustrates this decomposition by marking different components with different colours (see Appendix \tcr{B} for more details).
% (see Appendix B for visualization)

%This fine-grained decomposition disentangles heterogeneous error types, enabling precise credit assignment aligned with human perceptional judgement. 

\subsection{Component-Specific Rewarding}

We assign dedicated rewards to each functional component of a LaTeX sequence to achieve disambiguated rewarding. Formally, for a generated sequence $\mathbf{y}$ and its reference $\mathbf{y}^*$, we denote the component-specific rewards as
\begin{equation}
\begin{split}
\mathcal{R}(\mathbf{y}, \mathbf{y}^*) = \{ R_c \mid c \in \mathcal{C} \}, 
%&\qquad \qquad \quad 
%\\
%&\quad \mathcal{C} = \left\{\text{pkg}, \text{cap}, \text{struct}, \text{cell-app}, \text{align}, \text{vline}, \text{hline}\right\},
\end{split}
\end{equation}
where $R_c$ measures fidelity of component $c$. 
A strong LLM (\egno, GPT-4o) serves as an automatic judger (see Appendix \ref{app:C} for prompt) to evaluate each component. 
Each component reward is binary—1 if consistent with reference, 0 otherwise—providing clear, localized signals that disambiguate rewards across components.
%We use a strong LLM (\ieno, GPT-4o) as an automatic judger (see Appendix~\tcr{C} for prompt) to assess the consistency for each component.
%to score each functional component against the reference. For each component, the reward is binary: $1$ if consistent with the reference, $0$ otherwise (see Appendix~\tcr{C} for prompts). This component-wise feedback provides clear, localized signals, alleviating reward ambiguity.
% that alleviate reward ambiguity 

\subsection{Credit Assignment and Optimization}
With each functional component in $\mathcal{C}$ assigned a dedicated reward, we attribute credit only to the tokens corresponding to that component, avoiding cross-component interference.
%Each functional component in $\mathcal{C}$ is assigned a dedicated reward, and credit is attributed only to the corresponding tokens.
%We assign each functional component in $\mathcal{C}$ a dedicated reward and \textbf{attribute credit to only the corresponding tokens}. 
%This design prevents improvements in one component from being obscured by errors in others, leading to more reliable gradients and targeted policy updates.

As illustrated in Figure~\ref{fig:CSPO}, CSPO extends Group Relative Policy Optimization (GRPO) by augmenting the global reward with component-specific rewards.
Given a group of rollouts $\{\mathbf{y}^{(1)}, \ldots, \mathbf{y}^{(G)}\}$ sampled from the MLLM policy $\pi_\theta$, the \textbf{component-specific advantage} for component $c$ in rollout $g$ is defined as:
\begin{equation}
\begin{aligned}
A^{(g)}_{c} = \frac{R^{(g)}_{c} - \mu_{c}}{\sigma_{c} + \epsilon}, 
\end{aligned}
\label{eq:normalization}
\end{equation}
where $R^{(g)}_{c}$ represents the reward of component $c$ in rollout $g$, and $\mu_{c}$ and $\sigma_{c}$ denote the mean and standard deviation of the rewards for this component, respectively.

For the $t$-th token, we define the \textbf{token-level component advantage} w.r.t. component $c$ as
\begin{equation}
A^{(g)}_{c,t} =  A^{(g)}_{c} \cdot \mathbbm{I}[y_{t}^{(g)} \in c],
\label{eq:masking}
\end{equation}
where $\mathbbm{1}[\cdot]$ is an indicator function that activates only when the generated token $y_{t}^{(g)}$ belongs to component $c$. \textbf{This masking mechanism ensures that gradient updates are applied exclusively to relevant tokens, thereby facilitating precise component-specific optimization.}
%If this token does not belong to the component $c$, the advantage w.r.t. this component is zero and there is no gradient in optimization. 
%\in c$ denotes that the $t$-th token of $\mathbf{y}_g$ belongs to component $c$. 
%global consistency—\egno, component ordering mismatches or minor syntax errors. 

While $R^{(g)}_{c}$ optimizes component-wise fidelity, it may overlook inter-component dependencies. To compensate, we incorporate two global signals: the \emph{TEDS score} $R_{\text{TEDS}}$ for overall similarity, and a \emph{compile reward} $R_{\text{cmp}}$ to penalize non-compilable outputs. Their sum forms the global reward $R_{global}^{(g)} = R_{\text{TEDS}}^{(g)} + R_{\text{cmp}}^{(g)}$, whose normalized advantage is shared across all tokens within rollout $g$ as $A_{global,t}^{(g)} = A_{global}^{(g)}$. 
%. Similar to (\ref{eq:normalization}), we calculate the global advantage and denote it as $A_{global}^{(g)}$. Each token shares the same global advantage as 
For unified formulation, we extend the component set in (\ref{eq:C}) to include this global component as:
\begin{equation}
    \tilde{\mathcal{C}} = \mathcal{C} \cup \{\text{global}\}.
\end{equation} 

The CSPO objective maximizes the expected normalized advantage while regularizing the policy via a KL penalty to the reference model:
\begin{equation}
\begin{aligned}
\hspace*{-0.6em}
\mathcal{J}_{\text{CSPO}}(\theta) 
= \mathbb{E}_{(x,\mathbf{y}^*) \sim \mathcal{D}, \{\mathbf{y}^{(g)}\}_{g=1}^G \sim \pi_{\theta_{\text{old}}}(\cdot|x)} \Big[ \mathcal{L}(\theta)
- \\
\beta D_{\text{KL}}(\pi_{\theta} || \pi_{\text{ref}}) \Big], \, \text{where}\, \mathcal{L}(\theta)= \sum_{c \in \tilde{\mathcal{C}}}
\mathcal{L}_{c}(\theta).
\end{aligned}
\label{eq:CSPO}
\end{equation}
Each component-specific objective $\mathcal{L}_{c}(\theta)$ adopts a GRPO-style clipped surrogate loss:
\begin{equation}
\begin{split}
\hspace*{-0.6em}
\mathcal{L}_{c}(\theta) 
= &\frac{1}{G} \sum_{g=1}^{G} 
\frac{1}{|\mathbf{y}_c^{(g)}|}
\sum_{t=1}^{|\mathbf{y}_c^{(g)}|} 
\min \Big[
\rho_{g,t}(\theta) A_{c,t}^{(g)}, \, \\
&\text{clip}(\rho_{g,t}(\theta), 1-\epsilon, 1+\epsilon) A_{c,t}^{(g)}
\Big],
\end{split}
\label{eq:Lc}
\end{equation}
where $|\mathbf{y}^{(g)}_{c}|$ denotes the number of tokens associated to component $c$, and $\epsilon$ is the clipping threshold.

Based on (\ref{eq:CSPO})(\ref{eq:Lc}), the overall objective $\mathcal{L}(\theta)$ can be reformulated (derivation in Appendix \ref{app:D}) as 
\begin{equation}
\begin{split}
\hspace*{-0.6em}
\mathcal{L}(\theta) = &
\frac{1}{G} \sum_{g=1}^{G} \frac{1}{|\mathbf{y}^{(g)}|} \sum_{t=1}^{|\mathbf{y}^{(g)}|} 
\min \Big[ 
\rho_{g,t}(\theta) A^{(g)}_{t}, \, \\
&\text{clip}(\rho_{g,t}(\theta), 1-\epsilon, 1+\epsilon) A^{(g)}_{t} 
\Big],
\end{split}
\label{eq:L}
\end{equation}
where $A^{(g)}_{t}$ denotes the \textbf{token-level aggregated advantage}, which integrates the contributions of all components:
\begin{equation}
\begin{split}
    A^{(g)}_{t} = & \sum_{c \in \tilde{\mathcal{C}}} 
    \frac{|\mathbf{y}^{(g)}|}{|\mathbf{y}_c^{(g)}|} A^{(g)}_{c,t} \\
    = & \sum_{c \in \tilde{\mathcal{C}}} 
    \frac{|\mathbf{y}^{(g)}|}{|\mathbf{y}_c^{(g)}|} A^{(g)}_{c} \cdot \mathbbm{I}[y_{t}^{(g)} \in c].
    \label{eq:adv}
\end{split}    
\end{equation}
Here, $|\mathbf{y}^{(g)}|$ is the total token count of rollout $g$.
For the global component, $|\mathbf{y}_{global}^{(g)}| = |\mathbf{y}^{(g)}|$, whereas for others $|\mathbf{y}_{c}^{(g)}| < |\mathbf{y}^{(g)}|$. 

Notation summaries are in Appendix \ref{app:E}. Algorithm 1 summarizes the CSPO process. $w_c$ denotes weights for balancing the contributions of different components (see ablation in Appendix \ref{appsubsec:ablation}).

\begin{algorithm}[t]
\caption{Component-Specific Policy Optimization (CSPO)}
\label{alg:cspo}
\begin{algorithmic}[1]
\Require Pretrained policy $\pi_\theta$, reference policy $\pi_\text{ref}$, dataset $\mathcal{D}$, group size $G$, $w_c$
\For{each training iteration}
    \State Sample a batch of table images $x \sim \mathcal{D}$
    \State Generate $G$ rollouts $\{\mathbf{y}^{(g)}\}_{g=1}^G \!\sim\! \pi_\theta(\cdot|x)$
    \State Decompose each $\mathbf{y}^{(g)}$ into components $c \in \mathcal{C}$
    \State Compute component rewards $R_c^{(g)}$ and normalize: 
    $A_c^{(g)} = (R_c^{(g)} - \mu_c)/(\sigma_c + \epsilon)$, where $\epsilon = 1 \times 10^{-4}$
    \State Aggregate token-level advantages: 
    $A_t^{(g)} = \sum_{c \in \tilde{\mathcal{C}}} 
    \frac{|\mathbf{y}^{(g)}|}{|\mathbf{y}_c^{(g)}|} w_c 
    A_c^{(g)} \mathbbm{1}[y_t^{(g)} \in c]$
    \State Update $\pi_\theta$ by maximizing the clipped CSPO objective in (8).
\EndFor
\end{algorithmic}
\end{algorithm}

 % = \sum_{c \in \tilde{\mathcal{C}}} 
 %    \frac{|\mathbf{y}^{(g)}|}{|\mathbf{y}_c^{(g)}|}  A^{(g)}_{c} \cdot \mathbbm{I}[y_{t}^{(g)} \in c]
%= \sum_{c \in \tilde{\mathcal{C}}}     \frac{|\mathbf{y}^{(g)}|}{|\mathbf{y}_c^{(g)}|} A^{(g)}_{c} \cdot \mathbbm{I}[y_{t}^{(g)} \in c]
%, with $|\mathbf{y}_{global}^{(g)}| = |\mathbf{y}^{(g)}|$ and $|\mathbf{y}_{c}^{(g)}| < |\mathbf{y}^{(g)}|$ for other components. Notation summaries are in Appendix \tcr{E}.
% \begin{equation}
%     w_c \leftarrow \lambda  w_c + (1 - \lambda)\bar{w}_{c,i},
% \end{equation}
% where $\lambda \in [0,1)$ (default 0.9) controls temporal smoothness.

%\subsection{Training Algorithm}
%Algorithm~1 in Appendix \tcr{F} summarizes the CSPO process.
%The CSPO process is summarized in Algorithm 1 (Appendix \tcr{F}).

\section{Evaluation Metrics}
\label{subsec:evaluation}
To systematically assess model capabilities, we introduce a hierarchical evaluation framework that combines global similarity metrics—TEDS for overall matching and compile success rate—with newly proposed fine-grained diagnostics that disentangle structure, style, and content fidelity, enabling more interpretable analysis of model behavior beyond aggregated scores.
%To systematically assess the capabilities of models, we present a hierarchical evaluation framework: leveraging the global similarity metric TEDS for overall matching, compile successful rate (\ieno, renderability), supplemented by fine‑grained diagnostics across table structure, style, and content.
%To comprehensively assess both structural and functional fidelity, we report:
\begin{itemize}
    \item \textbf{Tree Edit Distance Similarity} \(TEDS\): measures the overall semantic and structural similarity between generated and reference LaTeX renderings (see Appendix \ref{app:A}).
    \item \textbf{Compilation Rate}  \(R\): the percentage of generated LaTeX codes that compile successfully.
    %\item \textbf{Component Accuracy}: precision of component-level correctness, computed via the same LLM evaluator used for reward computation.
\end{itemize}

\noindent\textbf{Fine-Grained Metrics.} 
%While TEDS and Compilation capture holistic quality, they do not diagnose specific strengths or weaknesses. 
To provide a more granular evaluation, we introduce fine-grained metrics automatically computed by an LLM (\egno, GPT-4o) that compares the predicted code $\mathbf{y}$ with the reference $\mathbf{y}^*$ (see Appendix \ref{app:C} for the detailed prompt). Each metric is defined at the \emph{table level}: a score of 1 indicates correctness, and 0 indicates at least one error. We evaluate along three main dimensions:
\begin{itemize}
  \item \textbf{Structural Correctness} ($S$): Verifies the consistency of structural elements, including merged cells (\texttt{\textbackslash multicolumn}, \texttt{\textbackslash multirow}) and cell positions.
  \item \textbf{Content Fidelity} ($C$): Checks equivalence of textual and numeric entries with the ground truth.
  \item \textbf{Stylistic Fidelity} ($Y$): Assesses presentation consistency, including line style ($Y_{\text{line}}$), column/cell alignment ($Y_{\text{align}}$)(\egno, left aligned), and cell formatting ($Y_{\text{cell}}$)(\egno, boldface, underline), where 
  $Y = Y_{\text{line}} \wedge Y_{\text{align}} \wedge Y_{\text{cell}}$. \(\wedge\) denotes logical AND operation.
\end{itemize}
We further define composite indicators to assess the overall fidelity in terms of structure, content,  style, and compilation success:
\begin{itemize}
  %\item \textbf{Information Fidelity} ($SC = S \wedge C$): Structural and content correctness.
  \item \textbf{Overall Fidelity} ($OF = S \wedge C \wedge Y  \wedge R$): Combined correctness across structure, content,  style, and compilation success.
\end{itemize}

\section{Experiments}

We conducted extensive experiments to validate the effectiveness of our proposed \textbf{CSPO} for table image-to-LaTeX generation. 
%Our goal is to examine (i) whether CSPO improves overall LaTeX generation fidelity, (ii) how component-specific optimization contributes to localized improvement, and \textcolor{red}{(iii) the role of adaptive reward balancing and global consistency in training stability.}

\subsection{Experimental Setup}

\noindent\textbf{Dataset.} We construct \emph{\benchmarkno}, a benchmark comprising 19000 pairs of table images and renderable LaTeX codes. In order to support complete table generation, each table code contains (i) necessary package declarations to ensure its compilation correctness; (ii) the caption of table\footnote{Existing datasets usually lack table captions and sample-wise compilation package specifications, which we include for table completeness.}; (iii) the body of the table that preserves the rich styles and formatting of the original table.  The dataset is curated from scientific articles collected from arXiv, where renderable table code is directly extracted from LaTeX sources. The articles span six major categories—Computer Science, Mathematics, Economics, Electrical Engineering and Systems Science, Quantitative Finance, and Statistics—covering publications from 2012 to 2025.
Each extracted LaTeX snippet is rendered into an image, resulting in tables with diverse aspect ratios, resolutions, and visual layouts.

We split the dataset to training set \emph{\benchmarkno-train} of 15000 samples and testing set \emph{\benchmarkno-test} of 4000 samples by a random partition. 
%To ensure compilation correctness, every code sample includes the necessary package declarations corresponding to the features used in the table. 
%Overall, \emph{\benchmarkno} provides a diverse and high-fidelity collection encompassing varied structural configurations, alignment schemes, and formatting styles, enabling comprehensive evaluation of model generalization and robustness.

We evaluate our method on three benchmarks: one \textbf{in-domain dataset}, \emph{\benchmarkno-test}, and two \textbf{out-of-domain datasets}, \ieno, \emph{DocGenome-table-1k} \citep{xia2024docgenome}, and \emph{Table2LaTeX-test-simple} \citep{ling2025table2latex}. (i)  \emph{DocGenome-table-1k} \citep{xia2024docgenome} is a 1,000-sample subset of DocGenome, where table images are automatically annotated by DocParser and cropped directly from raw PDFs. As a result, the dataset exhibits substantial variability in table localization accuracy, as well as occasional background clutter and partial table truncation, making it a significantly more challenging benchmark for robust table code generation.
%a 1,000-sample subset from DocGenome whose table images are automatically annotated by DocParser and cropped from original PDFs, introducing variability in resolution and table localization that makes this benchmark more challenging; 
(ii) \emph{Table2LaTeX-test-simple} consists of 496 test samples from \citep{ling2025table2latex}, where the table captions and colors are excluded during their dataset construction. 
%We follow its official evaluation protocol and report results on the simple split to complement the harder DocGenome setting.
%We evaluate the generalization capability of our model on datasets of (ii) and (iii). 
%(iii) \textbf{FinTabNet}~\cite{zhong2020image}, a large-scale dataset of financial tables annotated with structural markup.
%All LaTeX codes are normalized through rule-based postprocessing (e.g., removing redundant whitespaces, standardizing packages) to ensure consistent parsing. 
%A subset of the training data is reserved for validation and evaluator calibration.

\noindent\textbf{Evaluation Metrics.} Metrics defined in Section \ref{subsec:evaluation} are used for evaluation. By default, we evaluate the models on the fine-grained metrics by using GPT-4o as the judge. Consistent trends were observed when using other LLM judges (see Section~\ref{subsec:LLM_judge}).

\noindent\textbf{Implementation Details.}
We adopt vision–language models based on multimodal LLM backbones (\ieno, Qwen2.5-VL-3B, Qwen2.5-VL-7B) as base models, and train them using a two-stage procedure: supervised fine-tuning (SFT) followed by reinforcement learning (RL).
%We adopt a vision-language model from a multimodal LLM backbone (\ieno, \texttt{Qwen2.5-VL-3B}, \texttt{Qwen2.5-VL-7B}) as our base model. We conduct a two-stage training: SFT and RL.  
For SFT, we train on 10000 samples for one epoch, with an initial learning rate of 5e-6 and a batch size of 64. 
For RL, we use 5,000 training samples and train for two epochs, with a rollout batch size of 16 and four gradient accumulation steps.
%For GRPO, CSPO and VSGRPO, we use 5,000 samples and train for 2 epochs with a rollout batch size of 16 and four gradient-accumulation steps. 
We set the group size $G$ to 8, and the learning rate to 1e-6. 
We employ a fixed weighting scheme to balance the contributions of different components, \ieno, $\mathcal{L}(\theta)= \sum_{c \in \tilde{\mathcal{C}}} w_c \mathcal{L}_{c}(\theta)$, where we assign $w_{\text{global}} = 3$ to the global component and set $w_c = 1$ for each remaining component.
%All training is performed on 4×A100 GPUs. 
For evaluation, we generate a single rollout for each input image using greedy decoding.

\begin{table*}[th]
\centering
\resizebox{0.95\linewidth}{!}{
\begin{tabular}{l|cc|cccccc|c}
\toprule
Model & TEDS ↑ & OF ↑ & S ↑ & C ↑ & Y ↑ & \(Y_{line}\) ↑ & \(Y_{align}\) ↑ & \(Y_{cell}\) ↑ & R ↑ \\
\midrule
GPT-4o  \citep{openai2025gpt4o}                    & 72.3          & 6.8           & 65.0          & 73.1          & 9.4           & 23.0          & 54.0          & 72.5          & 99.9          \\
Gemini-2.5 Flash    \citep{comanici2025gemini}          & 66.2          & 11.1          & 78.8          & 79.2          & 12.9          & 33.5          & 47.2          & 81.3          & 99.3          \\
Qwen2.5-VL-72B    \citep{bai2025qwen2}            & 75.9          & 10.1          & 79.6          & 76.2          & 12.6          & 32.1          & 56.7          & 69.9          & 99.7          \\
%\midrule
Nougat \citep{blecher2023nougat} & 67.9 & 21.2 & 68.7 & 63.3 & 25.8 & 37.1 & 71.9 & 70.1 & 99.1 \\
\midrule
Qwen2.5-VL-3B             & 66.0          & 3.1           & 51.8          & 59.8          & 4.5           & 18.9          & 34.0          & 61.7          & 93.7          \\
Qwen2.5-VL-3B-VSGRPO$^*$  \citep{ling2025table2latex}      & 87.8          & 40.4          & 77.5          & 86.6          & 47.6          & 70.9          & 71.4          & 91.9          & \textbf{99.7} \\
Qwen2.5-VL-3B-GRPO        & 87.7          & 42.0          & 74.7          & 86.7          & 50.3          & 74.9          & 71.5          & 92.0          & 99.6          \\
\textbf{Qwen2.5-VL-3B-CSPO (Ours)} & \textbf{87.9} & \textbf{45.2} & \textbf{77.6} & \textbf{87.0} & \textbf{53.6} & \textbf{76.5} & \textbf{74.2} & \textbf{92.1} & 99.6          \\
\midrule
Qwen2.5-VL-7B             & 64.0          & 5.3           & 67.0          & 63.0          & 6.4           & 19.5          & 46.7          & 60.7          & 77.2          \\
Qwen2.5-VL-7B-VSGRPO$^*$  \citep{ling2025table2latex}      & 89.5          & 51.1          & 81.3          & 89.9          & 58.2          & 76.8          & 79.5          & 93.9          & 99.9          \\
Qwen2.5-VL-7B-GRPO        & 89.7          & 50.6          & 80.7          & 89.7          & 58.9          & \textbf{76.8} & 80.5          & 93.7          & \textbf{99.9} \\
\textbf{Qwen2.5-VL-7B-CSPO (Ours)} & \textbf{89.7} & \textbf{53.0} & \textbf{81.4} & \textbf{90.0} & \textbf{60.6} & 76.6          & \textbf{82.7} & \textbf{93.9} & 99.8  \\
\bottomrule
\end{tabular}   
}
\caption{
Performance comparisons on \emph{\benchmarkno-test} (with fine-grained metrics evaluated by GPT-4o). 
We report metrics across hierarchical dimensions. \textbf{Global metrics}: TEDS, Overall Fidelity (OF), Compilation Rate (R); \textbf{Fine-grained metrics}:   
(i) Structure Fidelity (S), (ii) Content Fidelity (C); 
(iii) Style Fidelity(Y): Line Style ($Y_{\text{line}}$), Alignment ($Y_{\text{align}}$), Cell Style ($Y_{\text{cell}}$). 
Higher scores (↑) indicate better performance. Note that all evaluation metrics, initially defined in the \([0,1]\) range (\egno, scoring and correctness measures), are presented as percentages (\%) in all the tables for clarity. $^*$ denotes that, for fair comparison, we reimplement the reward design of Ling \emph{et al.}~\cite{ling2025table2latex} within our codebase and dataset.}
%$^*$ denotes that we implemented the reward design from Ling \emph{et al.}~\cite{ling2025table2latex} for fair comparison.}
\label{tab:SOTA}
\end{table*}

\begin{table*}[th]
\centering
\resizebox{0.95\linewidth}{!}{
\begin{tabular}{l|ccccc|ccccc}
\toprule
& \multicolumn{5}{c}{\emph{DocGenome-table-1k}}                                         & \multicolumn{5}{c}{\emph{Table2LaTeX-test-simple}}                                        \\
Method                                                             & TEDS ↑        & OF ↑          & S ↑           & C ↑           & Y ↑           & TEDS ↑        & OF ↑          & S ↑           & C ↑           & Y ↑           \\
\midrule
GPT-4o  \citep{openai2025gpt4o}                    & 60.2          & 5.1           & 55.0          & 41.2          & 8.4           & 70.6          & 3.4           & 60.9          & 66.1          & 4.6           \\
Gemini-2.5 Flash    \citep{comanici2025gemini}          & 60.5          & 8.0           & 64.9          & 57.1          & 12.5          & 71.5          & 10.5          & 72.0          & 78.4          & 12.9          \\
Qwen2.5-VL-72B    \citep{bai2025qwen2}            & 60.3          & 5.0           & 61.2          & 53.9          & 8.2           & 74.7          & 7.5           & 64.9          & 66.5          & 11.7          \\
Nougat \citep{blecher2023nougat} & 30.8 & 0.8 & 27.7 & 4.4 & 8.8 & 27.0 & 0.4 & 20.6 & 2.8 & 5.7 \\
\midrule
Qwen2.5-VL-3B             & 50.2          & 1.1           & 43.5          & 37.0          & 1.7           & 63.5          & 2.0           & 39.1          & 44.2          & 3.0           \\
Qwen2.5-VL-3B-VSGRPO$^*$  \citep{ling2025table2latex}      & 71.5          & 17.8          & 60.3          & 53.8          & 30.9          & 81.5          & 22.6          & 63.3          & 74.2          & 30.7          \\
Qwen2.5-VL-3B-GRPO        & 71.3          & 18.9          & 60.0          & 54.4          & 33.1          & 81.5          & 24.4          & 58.3          & \textbf{75.6} & 32.1          \\
\textbf{Qwen2.5-VL-3B-CSPO (Ours)} & \textbf{72.5} & \textbf{21.0} & \textbf{63.6} & \textbf{54.7} & \textbf{34.5} & \textbf{82.3} & \textbf{26.0} & \textbf{64.3} & 74.4          & \textbf{33.9} \\
\midrule
Qwen2.5-VL-7B             & 56.6          & 3.9           & 51.2          & 46.3          & 6.0           & 66.4          & 4.0           & 54.6          & 58.7          & 5.4           \\
Qwen2.5-VL-7B-VSGRPO$^*$  \citep{ling2025table2latex}      & 73.9          & 25.9          & 67.8          & 63.0          & 39.6          & 82.9          & 33.9          & 66.9          & \textbf{80.2} & 42.3          \\
Qwen2.5-VL-7B-GRPO        & 74.7          & 26.9          & 68.5          & 62.7          & 38.7          & 83.1      & 34.7          & 64.7          & 79.8          & 44.8          \\
\textbf{Qwen2.5-VL-7B-CSPO (Ours)} & \textbf{74.7} & \textbf{29.2} & \textbf{68.8} & \textbf{63.6} & \textbf{41.1} & \textbf{83.5}          & \textbf{37.1} & \textbf{69.0} & 78.4          & \textbf{46.0} \\
\bottomrule
\end{tabular}
}
\caption{Generalization performance on \emph{DocGenome-table-1k} and \emph{Table2LaTeX-test-simple} (with fine-grained metrics evaluated by GPT-4o). $^*$ denotes that, for fair comparison, we reimplement the reward design of Ling \emph{et al.}~\cite{ling2025table2latex} within our codebase and dataset.}
\label{tab:generalization}
\end{table*}

\subsection{Quantitative Results}

We compare our method with representative baselines: \textbf{(i) closed-source multimodal large language models (MLLMs)} (\egno, GPT-4o and Gemini-2.5 Flash); \textbf{(ii) open-source MLLMs} (\egno, Qwen2.5-VL-72B, Qwen2.5-VL-3B, Qwen2.5-VL-7B); \textbf{(iii) specialized expert model} Nougat~\citep{blecher2023nougat}, which is an open-source system for LaTeX code conversion; \textbf{(iv)} for fairness, we compare  \textbf{baseline models} Qwen2.5-VL-3B/7B-GRPO (trained with SFT, and GRPO \cite{shao2024deepseekmath} using only our global reward $R_{global}$); In addition, we implemented the global reward design from Ling \emph{et al.}~\cite{ling2025table2latex} which aggregates code structure consistency and visual fidelity as a single reward, which we refer to as Qwen2.5-VL-7B-VSGRPO~\citep{ling2025table2latex}. Note that these models all suffer from reward ambiguity during RL.

\noindent\textbf{Main Result.}
% Table~\ref{tab:SOTA} reports the main results on \emph{\benchmarkno-test}. We compare our method with representative baselines. For general-purpose multimodal large language models (MLLMs), we include the closed-source GPT-4o and Gemini-2.5 Flash, as well as the open-source Qwen2.5-VL-72B. For specialized expert models, we evaluate against Nougat~\citep{blecher2023nougat}, an open-source system for LaTeX code conversion. In addition, we compare models trained under the same settings using VSGRPO and GRPO, respectively.
Table~\ref{tab:SOTA} reports the results on \emph{\benchmarkno-test}. We have five main observations/conclusions. \textbf{(i)} Our models Qwen2.5-VL-3B/7B-CSPO achieves the highest overall performance in terms of Overall Fidelity (OF) and TEDS scores, outperforming general-purpose MLLMs and baseline models. \textbf{(ii)} Under the same settings, CSPO consistently outperforms GRPO by 3.2\%/2.4\% on the 3B/7B models in terms of Overall Fidelity (OF), outperforms VSGRPO by 4.8\%/1.9\% on the 3B/7B models, demonstrating the effectiveness of our component-specific policy optimization. \textbf{(iii)} CSPO shows consistent improvement across structure (S), content (C), and style  fidelity (S), indicating that component-specific optimization effectively alleviates reward ambiguity and drives targeted fidelity enhancement. \textbf{(iv)} Qwen2.5-VL-7B-CSPO, with increased model capacity, achieves higher performance than Qwen2.5-VL-3B-CSPO. \textbf{(v)} Our fine-grained metrics enable a more diagnostic evaluation than TEDS, which only provides an aggregated score. They reveal that general-purpose MLLMs perform well on structure (S) and content (C) but lag behind on style fidelity (Y), while table-specialized models exhibit more balanced performance. We hope these metrics provide useful signals for guiding future table-to-LaTeX model design and optimization.

\noindent\textbf{Generalization Performance.} Table~\ref{tab:generalization} shows the comparisons on two out-of-domain datasets. 
On DocGenome-table-1k, CSPO achieves the best performance across the 3B and 7B models, delivering consistent gains in structure, content, and style fidelity. Compared to GRPO, Qwen2.5-VL-3B-CSPO and Qwen2.5-VL-7B-CSPO improve Overall Fidelity by 2.1\% and 2.3\%, respectively.
%, indicating robustness to resolution variations and table localization shifts introduced by automatic cropping.
On Table2LaTeX-test-simple, CSPO consistently outperforms GRPO/VSGRPO in terms of Overall Fidelity and TEDS. 
Overall, the results confirm the effectiveness of CSPO on out-of-domain datasets and model sizes.

\subsection{Ablation Studies}

Table~\ref{tab:ablation} shows the ablation results to validate the effectiveness of our designs on top of 3B base model (Qwen2.5-VL-3B). 
% In Table~\ref{tab:ablation}, we compare CSPO with several representative methods:
% \begin{itemize}
%     \item \textbf{SFT}: Supervised fine-tuning on LaTeX sequences without RL.
%     \item \textbf{CSPO (ours)}: Our full model with component-specific rewards and adaptive balancing.
%     \item \textbf{CSPO w/o Adaptive}: A variant without adaptive reward weighting.
%     %\item \textbf{CSPO w/o Components}: A variant excluding the global reward terms $R_{\text{TEDS}}$ and $R_{\text{cmp}}$.
%     \item \textbf{CSPO w/o Global}: A variant excluding the global reward terms $R_{\text{TEDS}}$ and $R_{\text{cmp}}$.
%     \item \textbf{CSPO w/o Components (\ieno, \textbf{GRPO})}: A variant excluding the component-specific rewarding and optimization, which degrades to  Standard Group Relative Policy Optimization~\cite{shao2024deepseekmath}. 
%     \item \textbf{CSPO w/ Comp. Sum}: A variant using the sum of the global reward and the component-specific rewards as a single reward for GRPO. 
% \end{itemize}

\begin{table}[t]
\centering
\resizebox{0.97\linewidth}{!}{
\begin{tabular}{l|cc|ccc}
\toprule
Model & TEDS ↑ & OF ↑ & S ↑ & C ↑ & Y ↑ \\
\midrule
Base                  & 66.0          & 3.1           & 51.8          & 59.8          & 4.5           \\
SFT                   & 87.1          & 39.8          & 75.8          & 86.1          & 47.3          \\
\midrule
GRPO                  & 87.7          & 42.0          & 74.7          & 86.7          & 50.3          \\
CSPO w/ Comp. Sum      & 87.7          & 42.2          & 76.3          & 87.0          & 50.9          \\
CSPO w/o Global       & 87.7          & 44.7          & \textbf{77.9} & 86.3          & 51.2          \\
\midrule
CSPO (Ours)           & \textbf{87.9} & \textbf{45.2} & 77.6          & \textbf{87.0} & \textbf{53.6} \\
\bottomrule
\end{tabular}
}
\caption{Ablation studies on 3B models evaluated on \emph{\benchmarkno-test}.}
\label{tab:ablation}
%\vspace{-3mm}
\end{table}

\begin{table*}[t]
\centering
%\resizebox{0.9\linewidth}{!}{
\resizebox{0.93\linewidth}{!}{
\begin{tabular}{l|l|cccc}
\toprule
LLM   Evaluators/Judge & Method & \multicolumn{1}{l}{Overall   Fidelity} & \multicolumn{1}{l}{Structure   Fidelity} & \multicolumn{1}{l}{Content   Fidelity} & \multicolumn{1}{l}{Style   Fidelity} \\
\midrule
\multirow{3}{*}{GPT-4o (by default)} & SFT & 39.8 & 75.8 & 86.1 & 47.3 \\
 & GRPO & 42.0 & 74.7 & 86.7 & 50.3 \\
 & CSPO (Ours) & \textbf{45.2} & \textbf{77.6} & \textbf{87.0} & \textbf{53.6} \\
\midrule
\multirow{3}{*}{Qwen3-Next-80b} & SFT & 37.6 & 72.0 & 88.4 & 47.2 \\
 & GRPO & 39.7 & 71.6 & 88.2 & 49.9 \\
 & CSPO (Ours) & \textbf{43.3} & \textbf{75.5} & \textbf{88.5} & \textbf{52.5}\\ 
 \midrule
 \multirow{3}{*}{DeepSeek-v3.2} & SFT & 34.0 & 75.7 & 86.5 & 42.0 \\
 & GRPO & 35.8 & 75.8 & \textbf{87.5} & 43.3 \\
 & CSPO (Ours) & \textbf{39.7} & \textbf{78.5} & 86.7 & \textbf{47.3}\\
 \midrule
 \multirow{3}{*}{GPT-5.2} & SFT & 30.6 & 66.1 & 75.7 & 38.7 \\
 & GRPO & 31.1 & 65.4 & 75.4 & 40.2 \\
 & CSPO (Ours) & \textbf{34.5} & \textbf{70.1} & \textbf{76.2} & \textbf{43.0} \\

\bottomrule
\end{tabular}
}
\caption{Performance comparisons among 3B SFT, GRPO and our CSPO models, with fine-grained metrics evaluated by using different LLM judges.}
\label{tab:multi-judge-eval-3B}
\end{table*}

\noindent\textbf{Effect of Reward Ambiguous.} Beyond comparing {CSPO} and {GRPO}, we further evaluate the variant {CSPO w/ Comp. Sum}, which naively aggregates the global reward and component-specific rewards into a single scalar reward optimized via GRPO.
{CSPO} outperforms {CSPO w/ Comp. Sum} by 3\% in OF, demonstrating that the performance gain stems from component-specific credit assignment and targeted optimization, rather than merely incorporating additional reward signals.
%Beside comparing {CSPO} and {GRPO}, we also compare with the scheme {CSPO w/ Comp. Sum}, which is a variant using the sum of the global reward and the component-specific rewards as a single reward for GRPO. {CSPO} outperforms {CSPO w/ Comp. Sum} by 3\% in OF, demonstrating that the benefit comes from the component-specific credit attribution and optimization, instead of the incorporation of more rewards.

\noindent\textbf{Effect of Global Rewards.}
Removing the global reward $R_{\text{global}}$ (\ieno, {CSPO w/o Global}), \ieno, $w_{global}=0$ leads to 0.5\% performance drops. 
This highlights the value of incorporating global constraints on overall context.
%This highlights the necessity of incorporating global-level constraints for structural coherence.

\noindent\textbf{SFT vs. RL.} SFT can effectively warm up the training, which quickly boosts the performance of the base model from 3.1\% to 39.8\% in terms of OF. RL with GRPO and our CSPO further improve the model capabilities by 2.2\% and 5.4\%, respectively.
%We can see that SFT only is inferior to our final model, especially on the out-of-domain datasets, where the generalization capability instead of simple memorization is required.     
% \noindent\textbf{Sensitivity to Group Size.}
% We further vary $G \in \{4,8,16\}$ and observe stable performance for $G \ge 8$, validating that CSPO benefits from moderate intra-group diversity without excessive sampling.

% \begin{table}[th]
% \centering
% \resizebox{0.9\linewidth}{!}{
% \begin{tabular}{l|cc|ccc}
% \toprule
% \textbf{Variant} & \textbf{TEDS ↑} & \textbf{OF ↑} & \textbf{S ↑} & \textbf{C ↑} & \textbf{Y ↑} \\
% \midrule
% GRPO (no components)      & 87.7 & 42.0 & 74.7 & 86.7 & 50.3 \\
% w/o Content Reward        & 87.9 & 45.2 & 77.4 & 86.9 & \textbf{53.4} \\
% w/o Structure Reward      & 87.8 & 43.3 & 77.0 & 86.0 & 51.8 \\
% w/o Style Reward          & 87.7 & 43.1 & \textbf{79.0} & \textbf{87.5} & 49.6 \\
% \midrule
% CSPO (Ours)               & \textbf{87.9} & \textbf{45.2} & 77.6 & 87.0 & \textbf{53.6} \\
% \bottomrule
% \end{tabular}
% }
% \caption{Ablation study on reward components on 3B models.}
% \label{tab:ablation-comp}
% \end{table}

\noindent\textbf{Effect of Specific Components.} We study the influence of different component rewards. Table~\ref{tab:ablation-comp} in Appendix \ref{appsubsec:ablation} shows that removing structure-related or style-related rewards leads to a significant performance drop, while the impact of content-related reward is comparatively smaller.
%\noindent\textbf{Effect of Specific Components.} We study the influence of different component rewards. Table~\ref{tab:ablation-comp} shows that removing structure-related or style-related rewards leads to a significant performance drop, while the impact of content-related reward is comparatively smaller. This is likely because content fidelity is already relatively high compared to structure and style (see GRPO performance), leaving less room for improvement. Removing all component-specific rewards reduces the model to the GRPO baseline, highlighting the importance of decomposed optimization.

%Removing all component-specific rewards reduces the model to the GRPO baseline, highlighting the importance of decomposed optimization.
%The \emph{content} reward (\ieno, caption and cell-appearance\footnote{In a cell, we do not disentangle the textual content and style in considering a cell is already a small unit in a table.}) primarily affects textual accuracy, the \emph{structure} reward (\ieno, table layout) determines structure accuracy, the \emph{style} reward (\ieno, column alignment and line style) influences style consistency. 
%, and the package reward influences the compilation rate
%Removing all the specific components reduces the model to the GRPO baseline, confirming the necessity of decomposed optimization.

\noindent\textbf{Effect of Weight $w_c$.} Appendix \ref{appsubsec:ablation} (Table~\ref{tab:ablation-wc}) shows the ablation study on component weight $w_c$. 

\noindent\textbf{Effect of Reward Granularity.} See Appendix \ref{appsubsec:ablation} for the ablation study on the impact of reward granularity.

%Figure~\ref{fig:sample-cellstyle} shows that CSPO recovers the cell style (marked by ellipse) accurately.
%Qwen2.5-VL-7B-GRPO incorrectly distorted the table structure and lines, leading to semantic corruption, whereas Qwen2.5-VL-7B-CSPO  better reproduced the fine-grained layout (merged cells) and stylistic consistency (lines), maintaining information consistency.
% GRPO错误展现了表格的结构和线型，造成信息损坏，而CSPO 更好地再现了精细的布局（合并单元格）和风格一致性（线条），保持了语义一致性。
%定性检查还确认了改进的编译成功率和语义对齐。

% \begin{figure}[t]
%     \centering
%     %\includegraphics[width=\linewidth]{figures/qualitative_comparison.pdf}
%     \caption{Qualitative comparisons between GRPO and CSPO. 
%     CSPO achieves superior structural and stylistic fidelity.}
%     \label{fig:qualitative}
% \end{figure}

\subsection{Reliability of LLM Evaluation}
\label{subsec:LLM_judge}

To ensure the robustness of LLM-based evaluation, we validate it from three aspects. 

First, we employ multiple independent LLM judges (\ieno, GPT-4o \citep{openai2025gpt4o}, Qwen3-Next-80B \citep{yang2025qwen3}, DeepSeek-v3.2 \citep{liu2025deepseek}, and GPT-5.2 \citep{openai2025gpt52}) in testing. Table~\ref{tab:multi-judge-eval-3B} show that the overall performance trends are largely consistent across different LLM judges for 3B models (see Appendix~\ref{appsubsec:llm-eval-reliability} for 7B models). This also demonstrates the effectiveness of our method. 

Second, we verify strong agreement between LLM judgments and human evaluation (for GPT-4o, approximately 90\% consistency on 500 randomly sampled examples). 

Third, we repeat the GPT-4o evaluation eight times and observe low variance (0.1–0.4) for metrics. Detailed analyses are provided in Appendix~\ref{appsubsec:llm-eval-reliability}.

\subsection{Qualitative Results}
\begin{figure}[t]
    \centering
    \includegraphics[width=1\linewidth]{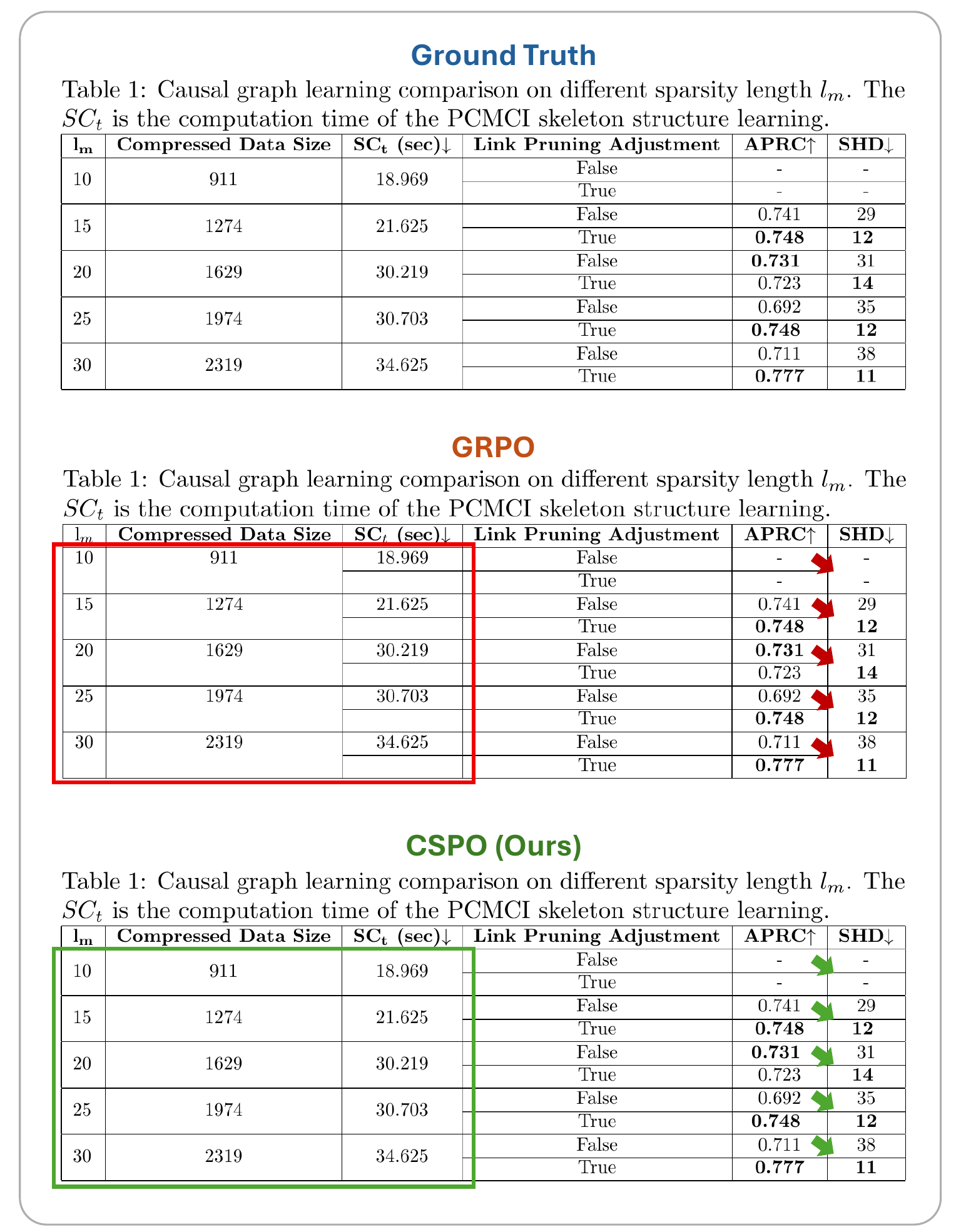}
    \caption{A typical example comparing GRPO and CSPO of 7B models, showing CSPO mitigates \textbf{structure} and \textbf{line style errors}.}
    \label{fig:sample}
    %\vspace{-3mm}
\end{figure}

We visualize the rendered table images from different models.
Figure~\ref{fig:sample} shows that our CSPO generates correct structure (marked by green box), line style (marked by green arrow), while GRPO suffers on structure (marked by red box) and line style (marked by red arrow). Figure~\ref{fig:sample-cellstyle} shows that CSPO recovers the cell style (marked by ellipse) accurately. See Appendix \ref{appsubsec:visualization} for more visualization. 

\begin{figure}[t]
    \centering
    \includegraphics[width=1\linewidth]{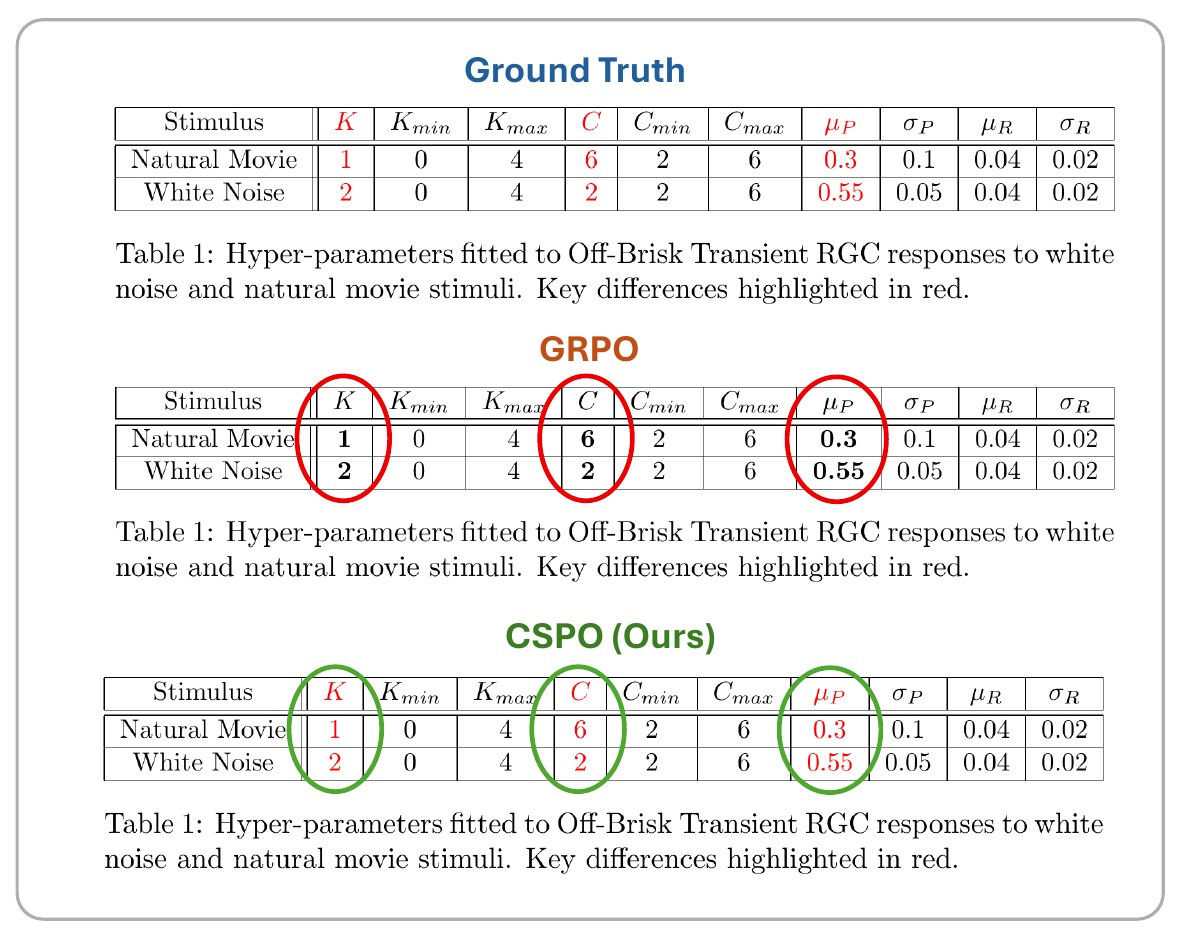}
    \caption{A typical example comparing GRPO and CSPO of 3B models, showing CSPO mitigates \textbf{cell style errors}.}
    \label{fig:sample-cellstyle}
    %\vspace{-3mm}
\end{figure}

\section{Conclusion}
We propose Component-Specific Policy Optimization (CSPO), a reinforcement learning framework that alleviates reward ambiguity in table image-to-LaTeX generation through component-specific rewarding, explicit credit assignment, and targeted policy optimization. Extensive experiments on both in-domain and out-of-domain benchmarks demonstrate that CSPO consistently improves structural, style, and content fidelity, highlighting the importance of addressing reward ambiguity in structured sequence generation. 

\section{Data Consent}
We collect data from arXiv, using only papers with CC BY, CC BY-SA, CC0, and CC BY-NC licenses. We will ensure that all collected data is used solely for research purposes, respecting the terms of the respective licenses. No personal or sensitive information is included, and all experiments and model training strictly follow ethical guidelines and data usage policies.

\section{Limitations}

Our CSPO design has alleviated the reward ambiguous problems and significantly enhanced the performance. However, there are still some limitations. (i) First, the overall fidelity of our models leaves room for further improvement (e.g., 53\% for the 7B-CSPO model). (ii) The scale of our training dataset is still small (\ieno, 15000 samples). More training data would further improve the performance.  (iii) CSPO relies on LLM-based evaluation during training, which introduces additional cost.
\bibliography{custom}

\appendix

% \section{Example Appendix}
% \label{sec:appendix}

% This is an appendix.

\clearpage

\section{TEDS}
\label{app:A}

\noindent\textbf{Global Similarity Metric} \(\mathrm{TEDS}\). Inspired by \cite{zhong2020image}, we adopt the Tree‑Edit‑Distance‑based Similarity (TEDS) score to measure the overall similarity. Particularly, we represent the generated LaTeX code and the ground‑truth code as rooted tree structures \(T_\mathrm{pred}\) and \(T_\mathrm{gt}\), respectively. The root has four types of children: \emph{table caption}, \emph{tabular} (which represents the column alignment manners and column lines), \emph{row entity}, and \emph{line entity}. The \emph{tabular} and \emph{row entity} nodes have multiple leaves to elaborate the table details. For example, under each \emph{row entity}, each cell corresponds to a leaf node. We compute a normalized edit distance:
\begin{equation}
\mathrm{TEDS}(T_\mathrm{pred}, T_\mathrm{gt}) = 1 - \frac{\mathrm{EditDist}(T_\mathrm{pred}, T_\mathrm{gt})}{\max(|T_\mathrm{pred}|,\;|T_\mathrm{gt}|)},
\end{equation}
where \(\mathrm{EditDist}\) denotes tree-edit distance, and \(|T|\) denotes the number of nodes in \(T\). The cost of insertion, deletion, editing are all 1. Note that we do not take the package headers into consideration in the TEDS measure.

\section{Component Decomposition}
\label{app:B}

The components in $\mathcal{C}$ are defined as follows:
\begin{itemize}
    \item \textbf{pkg}: imported packages (\egno, \texttt{booktabs}, \texttt{multirow});
    \item \textbf{struct}: tabular structure, including row/column merges and overall layout consistency;
    \item \textbf{cap}: table caption consistency;
    \item \textbf{cell-app}: cell-level appearance, covering textual fidelity and formatting consistency (\egno, bold, underline);
    \item \textbf{align}: column alignment type, specifying whether each column is centered, left-, or right-aligned (\texttt{c}, \texttt{l}, \texttt{r});
    %column alignment (\texttt{c}, \texttt{l}, \texttt{r});
    \item \textbf{vline}: vertical rule placement (\texttt{|});
    \item \textbf{hline}: horizontal rule placement (\verb|\hline|, \verb|\cline|).
\end{itemize}

%\tcr{show more examples here ...}

\section{Prompts for Rewarding and Evaluation}
\label{app:C}
\subsection{Prompt for Component Rewarding}
\begin{figure}[t]
\centering
\includegraphics[width=1\linewidth]{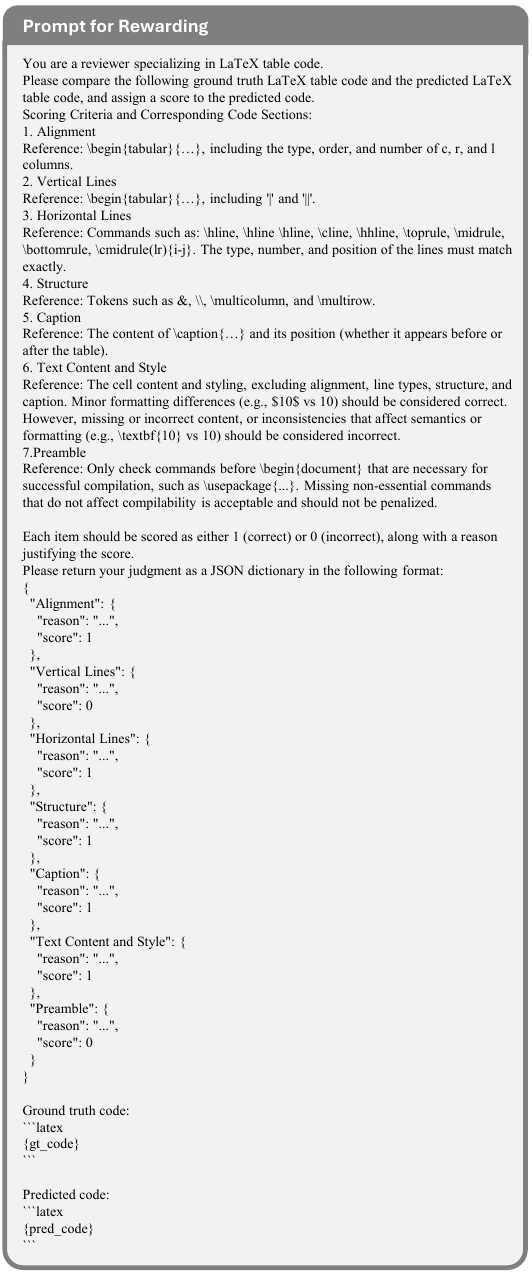}
\caption{Prompt for LLM-based rewarding for components of package dependencies, caption correctness, structural organization, cell appearance, column alignment, and rule placement (vline, hline).}
\label{fig:prompt-reward}
\end{figure}

To automatically provide reward for each component, we leverage a strong LLM as the reward model, \ieno, an automatic evaluator, to compare the predicted code $\mathbf{y}$ against the ground-truth $\mathbf{y}^*$ to score each component. For each component, the evaluator checks consistency between prediction and reference. If the component matches, the reward is set to $1$; otherwise, if inconsistencies are detected, the reward is set to $0$.
The detailed prompt is shown in Figure~\ref{fig:prompt-reward}.

\subsection{Prompt for Fine-grained Evaluation}
\begin{figure}[t]
\centering
\includegraphics[width=1\linewidth]{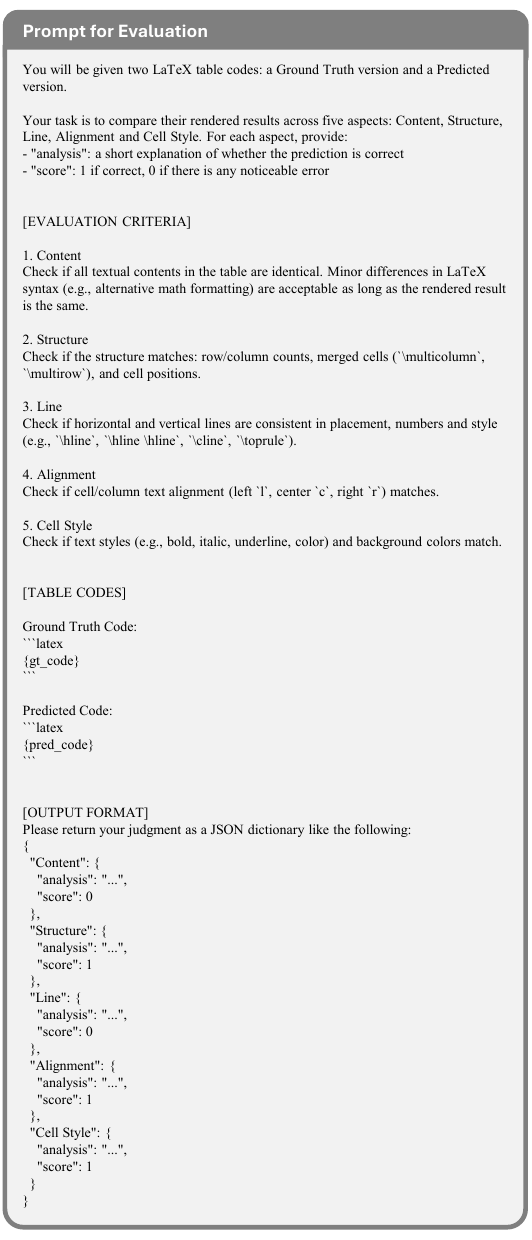}
\caption{Prompt for LLM-based fine-grained fidelity evaluation in terms of content, structure, and style (line style, alignment, and cell style).}
\label{fig:prompt-score}
\end{figure}

Figure~\ref{fig:prompt-score} provides the prompt for the fine-grained fidelity evaluation in terms of content, structure, and style (line style, alignment, and cell style).

%\section{Derivation of Aggregated Token-level Advantage}
\section{Derivation of Aggregated Objective in CSPO}
\label{app:D}

In this appendix, we provide a detailed derivation of Eq.~(10) from Eq.~(8) and (9).

\subsection{Component-Level Objective Recap}

Recall that CSPO decomposes the overall objective into component-specific terms:
\begin{equation}
\mathcal{L}(\theta) = \sum_{c \in \tilde{\mathcal{C}}} \mathcal{L}_c(\theta),
\end{equation}
where each $\mathcal{L}_c(\theta)$ corresponds to the clipped policy gradient surrogate for component $c$:
\begin{equation}
\begin{aligned}
\mathcal{L}_{c}(\theta) 
= \frac{1}{G} \sum_{g=1}^{G} 
\frac{1}{|\mathbf{y}_c^{(g)}|}
\sum_{t=1}^{|\mathbf{y}_c^{(g)}|} 
\min \Big[
\rho_{g,t}(\theta) A_{c,t}^{(g)}, \, \\
\text{clip}(\rho_{g,t}(\theta), 1-\epsilon, 1+\epsilon) A_{c,t}^{(g)}
\Big],
\end{aligned}
\label{eq:Lc_app}
\end{equation}
$|\mathbf{y}^{(g)}_{c}|$ indicates the number of tokens associated with component $c$ in rollout $\mathbf{y}^{(g)}$, $\epsilon$ denotes the clipping threshold.

\subsection{From Component-Level to Unified Objective}

To unify all component-level objectives into a single token-level surrogate, we first sum over all components:
\begin{equation}
%\begin{aligned}
\begin{split}
\hspace*{-0.6em}
\mathcal{L}(\theta)
= &\frac{1}{G} \sum_{g=1}^{G}
\sum_{c \in \tilde{\mathcal{C}}}
\frac{1}{|\mathbf{y}_c^{(g)}|}
\sum_{t=1}^{|\mathbf{y}_c^{(g)}|}
\min \Big[
\rho_{g,t}(\theta) A_{c,t}^{(g)}, \, \\
&\text{clip}(\rho_{g,t}(\theta), 1-\epsilon, 1+\epsilon) A_{c,t}^{(g)}
\Big].
%\end{aligned}
\end{split}
\label{eq:app_sum}
\end{equation}

Since each rollout $\mathbf{y}^{(g)}$ contains all tokens across components, we can reorganize the double sum over $(c, t)$ into a single sum over all token indices $t$ in $\mathbf{y}^{(g)}$. 
To ensure proper weighting across components with different token spans, we normalize by the total token length $|\mathbf{y}^{(g)}|$:
\begin{equation}
\begin{split}
\hspace*{-0.6em}
\mathcal{L}(\theta)
= \frac{1}{G} \sum_{g=1}^{G}
\sum_{c \in \tilde{\mathcal{C}}}
\frac{1}{|\mathbf{y}_c^{(g)}|} \frac{|\mathbf{y}^{(g)}|}{|\mathbf{y}^{(g)}|}
\sum_{t=1}^{|\mathbf{y}_c^{(g)}|}
\min \Big[
\rho_{g,t}(\theta) A_{c,t}^{(g)}, \, \\
\hspace*{-0.3em}\text{clip}(\rho_{g,t}(\theta), 1-\epsilon, 1+\epsilon) A_{c,t}^{(g)}
\Big]\\
= \frac{1}{G} \sum_{g=1}^{G}
\frac{1}{|\mathbf{y}^{(g)}|}
\sum_{c \in \tilde{\mathcal{C}}}\sum_{t=1}^{|\mathbf{y}_c^{(g)}|}  \frac{|\mathbf{y}^{(g)}|}{|\mathbf{y}_c^{(g)}|}
\min \Big[
\rho_{g,t}(\theta) A_{c,t}^{(g)}, \, \\
\text{clip}(\rho_{g,t}(\theta), 1-\epsilon, 1+\epsilon) A_{c,t}^{(g)}
\Big]\\
= \frac{1}{G} \sum_{g=1}^{G}
\frac{1}{|\mathbf{y}^{(g)}|}
\sum_{c \in \tilde{\mathcal{C}}}\sum_{t=1}^{\tcr{|\mathbf{y}^{(g)}|}}  \frac{|\mathbf{y}^{(g)}|}{|\mathbf{y}_c^{(g)}|}
\min \Big[
\rho_{g,t}(\theta) A_{c,t}^{(g)}, \, \\
\text{clip}(\rho_{g,t}(\theta), 1-\epsilon, 1+\epsilon) A_{c,t}^{(g)}
\Big]\\
= \frac{1}{G} \sum_{g=1}^{G}
\frac{1}{|\mathbf{y}^{(g)}|}
\sum_{t=1}^{|\mathbf{y}^{(g)}|} 
\min \Big[
\rho_{g,t}(\theta) A_{t}^{(g)}, \, \\
\text{clip}(\rho_{g,t}(\theta), 1-\epsilon, 1+\epsilon) A_{t}^{(g)}
\Big], 
\end{split}
\label{eq:app_L}
\end{equation}
where the aggregated token-wise advantage $A_t^{(g)}$ is as:
\begin{equation}
A_t^{(g)} = 
\sum_{c \in \tilde{\mathcal{C}}}
\frac{|\mathbf{y}^{(g)}|}{|\mathbf{y}_c^{(g)}|}
A_{c,t}^{(g)}.
\label{eq:app_adv_final}
\end{equation}
Note that the second- to third- equation holds because $A_{c,t}^{(g)} = 0$ whenever $y_{g,t} \notin c$.

%\subsection{Discussion}

This formulation ensures that:
\begin{itemize}
    \item \textbf{Balanced credit assignment:} Components with fewer associated tokens (e.g., global style indicators) are up-weighted, preventing their gradients from vanishing.
    % \item \textbf{Consistent normalization:} The scaling factor $|\mathbf{y}^{(g)}|^{-1}$ maintains comparable magnitude across rollouts of different lengths.
    \item \textbf{Unified training signal:} By aggregating all component-specific advantages into token-level surrogate, CSPO allows end-to-end optimization using a GRPO-style objective.
    %while preserving interpretability at the component level.
\end{itemize}

\section{Notation Summary}
\label{app:E}
The main notations used in the CSPO framework are summarized in Table~\ref{tab:notation}.

\begin{table}[t]
\centering
\caption{Notation summary for CSPO framework.}
\label{tab:notation}
\resizebox{1\linewidth}{!}{
\begin{tabular}{cl}
\toprule
\textbf{Notation} & \textbf{Description} \\
\midrule
$\mathbf{x}$ & Input table image \\
$\mathbf{y}$ & Generated LaTeX sequence \\
$\mathbf{y}^*$ & Reference LaTeX sequence \\
$\pi_\theta$ & Policy model with parameters $\theta$ \\
$T$ & Length of LaTeX sequence \\
$G$ & Number of rollouts per group \\
$\mathbf{y}^{(g)}$ & The $g$-th rollout in a group \\
$|\mathbf{y}^{(g)}|$ & Total token count in rollout $g$ \\
\midrule
$\mathcal{C}$ & Set of functional components \\
$\tilde{\mathcal{C}}$ & Extended component set including global \\
$c$ & A specific component in $\mathcal{C}$ or $\tilde{\mathcal{C}}$ \\
$|\mathbf{y}_c^{(g)}|$ & Number of tokens in component $c$ for rollout $g$ \\
\midrule
$R_c$ & Component-specific reward for component $c$ \\
$R^{(g)}_c$ & Reward of component $c$ in rollout $g$ \\
$R_{\text{global}}^{(g)}$ & Global reward (TEDS + compilation) for rollout $g$ \\
$A^{(g)}_c$ & Component-specific advantage for component $c$ \\
$A^{(g)}_{c,t}$ & Token-level component advantage at position $t$ \\
$A^{(g)}_t$ & Token-level aggregated advantage at position $t$ \\
\midrule
$w_c$ & Weight for component $c$ \\
$\alpha$ & Fixed weight for global component \\
$\beta$ & Sensitivity parameter for adaptive weighting \\
$\lambda$ & Smoothing parameter for weight updates \\
$\epsilon$ & Clipping threshold for PPO-style objective \\
\midrule
$\mathbbm{I}[\cdot]$ & Indicator function \\
$\rho_{g,t}(\theta)$ & Importance ratio $\pi_\theta(y_t^{(g)} | y_{<t}^{(g)}, \mathbf{x}) / \pi_{\theta_{\text{old}}}(y_t^{(g)} | y_{<t}^{(g)}, \mathbf{x})$ \\
\bottomrule
\end{tabular}
}
\end{table}

\begin{table*}[t]
\centering
\resizebox{0.90\linewidth}{!}{
\begin{tabular}{l|cc|cccccc|c}
\toprule
\textbf{Variant} & TEDS ↑ & OF ↑ & S ↑ & C ↑ & Y ↑ & \(Y_{line}\) ↑ & \(Y_{align}\) ↑ & \(Y_{cell}\) ↑ & R ↑ \\
\midrule
GRPO (no components)      & 87.7          & 42.0          & 74.7          & 86.7          & 50.3          & 74.9          & 71.5          & 92.0          & 99.6          \\
CSPO w/o Content Reward        & 87.9          & 45.2          & 77.4          & 86.9          & 53.4          & 76.0          & 74.4          & \textbf{92.5} & \textbf{99.7} \\
CSPO w/o Structure Reward      & 87.8          & 43.3          & 77.0          & 86.0          & 51.8          & 73.3          & \textbf{74.5} & 92.4          & 99.6          \\
CSPO w/o Style Reward          & 87.7          & 43.1          & \textbf{79.0} & \textbf{87.5} & 49.6          & 72.3          & 72.7          & 92.4          & 99.6          \\
\midrule
CSPO (Ours) & \textbf{87.9} & \textbf{45.2} & 77.6          & 87.0          & \textbf{53.6} & \textbf{76.5} & 74.2          & 92.1          & 99.6   \\
\bottomrule
\end{tabular}
}
\caption{Ablation study on reward components on 3B models. }
% CSPO with all components achieves the best balance across metrics. 
\label{tab:ablation-comp}
\end{table*}

\begin{table*}[t]
\centering
\resizebox{0.92\linewidth}{!}{
\begin{tabular}{l|cc|cccccc|c}
\toprule
\textbf{Variant} & TEDS ↑ & OF ↑ & S ↑ & C ↑ & Y ↑ & \(Y_{line}\) ↑ & \(Y_{align}\) ↑ & \(Y_{cell}\) ↑ & R ↑ \\
\midrule
GRPO ($w_{global}=10, w_{comp}=0$) & 87.7          & 42.0          & 74.7          & 86.7          & 50.3          & 74.9          & 71.5          & 92.0          & 99.6          \\
CSPO ($w_{global}=7, w_{comp}=3$)  & \textbf{88.0} & 43.6          & 76.4          & 85.9          & 52.1          & 74.4          & 74.1          & 91.9          & 99.8          \\
CSPO ($w_{global}=5, w_{comp}=5$)  & 87.9          & 44.8          & 77.1          & 86.9          & 53.1          & 76.0          & 73.8          & \textbf{92.7} & 99.7          \\
CSPO ($w_{global}=3, w_{comp}=7$)  & 87.9          & \textbf{45.2} & 77.6          & \textbf{87.0} & \textbf{53.6} & \textbf{76.5} & \textbf{74.2} & 92.1          & 99.6          \\
CSPO ($w_{global}=0, w_{comp}=10$) & 87.7          & 44.7          & \textbf{77.9} & 86.3          & 51.2          & 73.7          & 73.4          & 92.1          & \textbf{99.8}\\
\bottomrule
\end{tabular}
}
\caption{Ablation on reward weights for different components on 3B models.}
\label{tab:ablation-wc}
\end{table*}

\section{More Results}

\subsection{More Ablation}
\label{appsubsec:ablation}

\noindent\textbf{Effect of Specific Components.} We study the influence of different component rewards. Table~\ref{tab:ablation-comp} shows that removing structure-related or style-related rewards leads to a significant performance drop, while the impact of removing the content-related reward is comparatively smaller. This is likely because content fidelity is already relatively high compared to structure and style fidelity (see the performance of GRPO). As a result, there is less room for further optimization on content, and incorporating content-specific rewards yields a smaller marginal effect. Removing all component-specific rewards reduces the model to the GRPO baseline, highlighting the importance of decomposed optimization.

Note that here the \emph{content} reward (\ieno, caption and cell-appearance\footnote{In a cell, we do not disentangle the textual content and style in considering a cell is already a small unit in a table.}) primarily affects text accuracy, the \emph{structure} reward (\ieno, table layout) determines structure accuracy, the \emph{style} reward (\ieno, column alignment and line style) influences style consistency. 

Actually, the scheme CSPO w/o Style Reward only removes partial styles (\ieno, column alignment and line style) while still keeping the cell style. This is because in a cell, we cannot disentangle the textual content and style in considering a cell is already a small unit in a table. Then the cell content and style are still optimized. That may be the reason that there is no performance drop on style fidelity.
%, and the package reward influences the compilation rate

%Removing all the specific components reduces the model to the GRPO baseline, confirming the necessity of decomposed optimization.

%\noindent\textbf{Effect of Weight $w_c$.} Table~\ref{tab:ablation-wc} shows the ablation study on component weight $w_c$ at coarse granularity. Preserving both global rewards and component-specific rewards achieves the best performance.

\noindent\textbf{Effect of Reward Weights.}
Table~\ref{tab:ablation-wc} shows the ablation study in different reward weight configurations, where $w_{comp}$ denotes the sum of component-specific reward weights ($w_{comp} = \sum_{c \in \mathcal{C}} w_c$, with $w_c$ equal for different $c \in \mathcal{C}$). CSPO consistently outperforms GRPO across all configurations, indicating that the effectiveness of CSPO does not rely on a specific weight setting. Among all configurations, $w_{comp}=7$ achieves the best Overall Fidelity.

\begin{table}[t]
\centering
\resizebox{1.0\linewidth}{!}{
\begin{tabular}{l|ccccc}
\toprule
Method            & TEDS          & OF            & Structure     & Content       & Style         \\
\midrule
GRPO & 87.7          & 42.0          & 74.7          & 86.7          & 50.3 \\
\hline
CSPO (Graded 0-3) & \textbf{88.0}        & 44.4                 & 77.3                 & 86.7                 & 51.8                 \\
CSPO (Binary 0/1) & 87.9                 & \textbf{45.2}        & \textbf{77.6}        & \textbf{87.0}        & \textbf{53.6}        \\
\bottomrule
\end{tabular}
}
\caption{Performance comparison of CSPO using binary (0 or 1) and graded  (0 to 3) reward on 3B models.}
\label{tab:graded-reward}
\end{table}

\noindent\textbf{Effect of Reward Granularity.} In our CSPO, by default we employ binary component-specific rewards. We further investigate the impact of reward granularity by introducing a graded scoring scheme. Specifically, instead of assigning a binary signal (0 or 1), each component is evaluated on a four-level scale (0–3), where 3 indicates perfect alignment, 2 denotes minor yet interpretable gaps, 1 corresponds to major errors, and 0 represents failed or invalid outputs. As shown in Table~\ref{tab:graded-reward}, compared with graded component-specific reward, the binary component-specific reward scheme achieves comparable TEDS and a higher Overall Fidelity. Both the two schemes obviously outperforms the GRPO baseline and we use binary reward by default.
%, and consistently outperforms graded reward across fine-grained metrics, resulting in a higher Overall Fidelity. 
%These results suggest that, while graded rewards provide more expressive supervision, the simpler binary signals may lead to more stable and effective optimization in our setting.

%These results suggest that, while graded rewards provide more expressive supervision, the simpler binary signals may lead to more stable and effective optimization in our setting.

\begin{table*}[t]
\centering
\resizebox{0.93\linewidth}{!}{
\begin{tabular}{l|l|cccc}
\toprule
LLM   Evaluators/Judge & Method & \multicolumn{1}{l}{Overall   Fidelity} & \multicolumn{1}{l}{Structure   Fidelity} & \multicolumn{1}{l}{Content   Fidelity} & \multicolumn{1}{l}{Style   Fidelity} \\
\midrule
\multirow{3}{*}{GPT-4o (by default)} & SFT & 50.7  & 79.7 & 90.3 & 59.0 \\
& GRPO & 50.6 & 80.7 & 89.7 & 58.9 \\
& \textbf{CSPO (Ours)} & \textbf{53.0} & \textbf{81.4}  & \textbf{90.0} & \textbf{60.6} \\
\midrule
\multirow{3}{*}{Qwen3-Next-80b} & SFT & 47.1 & 77.1 & 91.5 & 57.5 \\
& GRPO & 49.7 & 77.6 & 91.5 & 59.8 \\
& \textbf{CSPO (Ours)} & \textbf{52.0} & \textbf{79.5} & \textbf{92.2} & \textbf{61.8} \\
\midrule
\multirow{3}{*}{DeepSeek-v3.2} & SFT & 43.3 & 79.3 & 90.4 & 51.4 \\
& GRPO & 45.8 & 80.5 & 90.3 & 54.7 \\
& \textbf{CSPO (Ours)} & \textbf{47.8} & \textbf{81.6} & \textbf{91.0} & \textbf{56.2} \\
\midrule
\multirow{3}{*}{GPT-5.2} & SFT & 38.7 & 70.0 & 80.8 & 48.0 \\
& GRPO  & 40.4 & 71.6 & 80.3 & 50.6 \\
& \textbf{CSPO (Ours)} & \textbf{42.6} & \textbf{73.8} & \textbf{81.8} & \textbf{51.8}  \\                     
\bottomrule
\end{tabular}
}
\caption{Performance comparisons among 7B SFT, GRPO and our CSPO models, with fine-grained metrics evaluated by using different LLM judges.}
\label{tab:multi-judge-eval-7B}
\end{table*}

\begin{table}[t]
\centering
\resizebox{1.0\linewidth}{!}{
\begin{tabular}{l|cccc}
\toprule
Evaluator                   & Precision & Recall & F1   & Accuracy \\
\midrule
GPT-4o                      & 89.8      & 86.9   & 88.3 & 89.8     \\
Qwen3-Next-80b & 88.8      & 85.6   & 87.2 & 88.8     \\
DeepSeek-v3.2               & 92.4      & 76.9   & 84.0 & 87.0     \\
GPT-5.2                     & 97.5      & 70.7   & 82.0 & 86.2  \\
\bottomrule
\end{tabular}
}
\caption{Agreement between LLM judges with human annotations.}
\label{tab:llm-human-agreement}
\end{table}

\begin{table}[t]
\centering
\resizebox{0.93\linewidth}{!}{
\begin{tabular}{l|cccc}
\toprule
Method                        & OF   & Structure & Content & Style \\
\midrule
GRPO            & 42.0 & 74.7      & 86.7    & 50.3  \\
CSPO (Reported) & 45.2 & 77.6      & 87.0    & 53.6  \\
CSPO (Mean ×8)  & 44.9 & 77.7      & 86.8    & 52.9  \\
CSPO (Var ×8)   & 0.3  & 0.1       & 0.1     & 0.4  \\
\bottomrule
\end{tabular}
}
\caption{Stability of GPT-4o evaluator in testing (8 independent runs) evaluated on 3B models.}
\vspace{-5mm}
\label{tab:llm-eval-stability}
\end{table}

\subsection{Reliability of LLM-based Evaluation}
\label{appsubsec:llm-eval-reliability}

In this work, we employ a strong LLM as an automatic judge to assess component-wise and overall fidelity. We use GPT-4o by default. Given concerns about evaluator bias and potential circularity, we conduct additional analyses on cross-model consistency, human alignment, and evaluation variance.

\noindent\textbf{Cross-LLM Consistency.} We evaluate model performance using multiple independent LLM-based judges in testing, including GPT-4o, Qwen3-Next-80B-A3B-Instruct, DeepSeek-v3.2, and GPT-5.2. Table~\ref{tab:multi-judge-eval-3B} in the main manuscript and Table~\ref{tab:multi-judge-eval-7B} here show that the overall performance trends are largely consistent across different LLM judges for 3B models and 7B models, respectively. In particular, CSPO consistently outperforms GRPO on most metrics. This indicates that our conclusions are not specific to a particular LLM judge, but instead reflect robust improvements in generation quality.

% \noindent\textbf{Cross-LLM Consistency.} We evaluate model performance using multiple independent LLM-based judges in testing, including GPT-4o, Qwen3-Next-80B-A3B-Instruct, DeepSeek-v3.2, and GPT-5.2. Table~\ref{tab:multi-judge-eval-3B} and Table~\ref{tab:multi-judge-eval-7B} show that the overall performance trends are largely consistent across different LLM judges for 3B models and 7B models, respectively. In particular, CSPO consistently outperforms GRPO on most metrics, with only a minor deviation on a single fine-grained metric. This indicates that our conclusions are not specific to a particular LLM judge, but instead reflect robust improvements in generation quality.

\noindent\textbf{Human–LLM Agreement.} We further assess the alignment between LLM-based judgments and human evaluation, where annotators label the overall fidelity (binary OF) of each output on a randomly sampled subset of 500 samples. Table~\ref{tab:llm-human-agreement} reports precision, recall, F1-score, and accuracy. GPT-4o and Qwen3-Next-80B-A3B-Instruct achieve strong agreement with human annotations, with F1-scores of 88.3 and 87.2, and accuracies of 89.8\% and 88.8\% (\ieno $\sim$90\%), respectively. DeepSeek-v3.2 and GPT-5.2 also show reasonable alignment, albeit with lower recall. 

Overall, these results suggest that LLM-based evaluation closely aligns with human judgment, supporting its use as a reliable proxy in structured table-to-LaTeX generation.
%These results indicate that LLM-based evaluation is consistent with human judgment in structured table-to-LaTeX generation.

\noindent\textbf{Evaluation Variance and Stability.} We assess the stability of LLM-based evaluation by repeating GPT-4o evaluation eight times on the same set of 4000 predictions from Qwen2.5-VL-3B-CSPO. As shown in Table~\ref{tab:llm-eval-stability}, the low variance (0.1–0.4) across all metrics indicates that the LLM-based judge produces stable evaluation signals.

\subsection{Training Dynamics}
\begin{figure*}[t]
    \centering
    \includegraphics[width=1\linewidth]{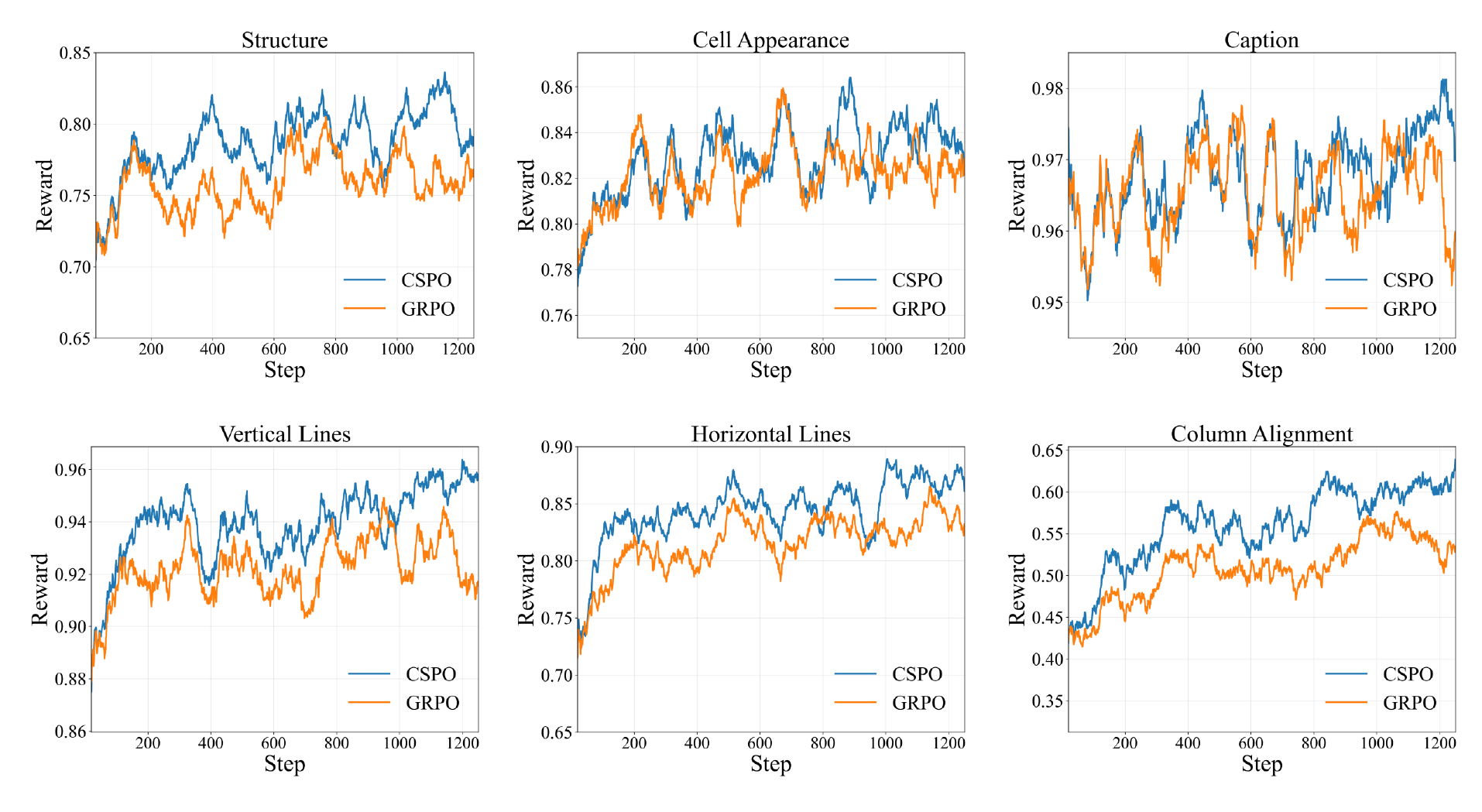}
    \caption{Training curves for 3B-GRPO and 3B-CSPO (ours). All curves are smoothed using a moving average (MA) for better visualization.}
    \label{fig:training_curves}
    %\vspace{-3mm}
\end{figure*}

We compare the training dynamics of CSPO with GRPO. Figure~\ref{fig:training_curves} presents the reward curves for different components during training. For GRPO, we report component-specific rewards, although these rewards do not influence its optimization. Compared with GRPO, CSPO converges to higher rewards on Structure, Lines, Alignment, and Caption, while reaching a similar level for Cell Appearance.

\subsection{More Visualization}
\label{appsubsec:visualization}

\begin{figure}[t]
    \centering
    \includegraphics[width=0.95\linewidth]{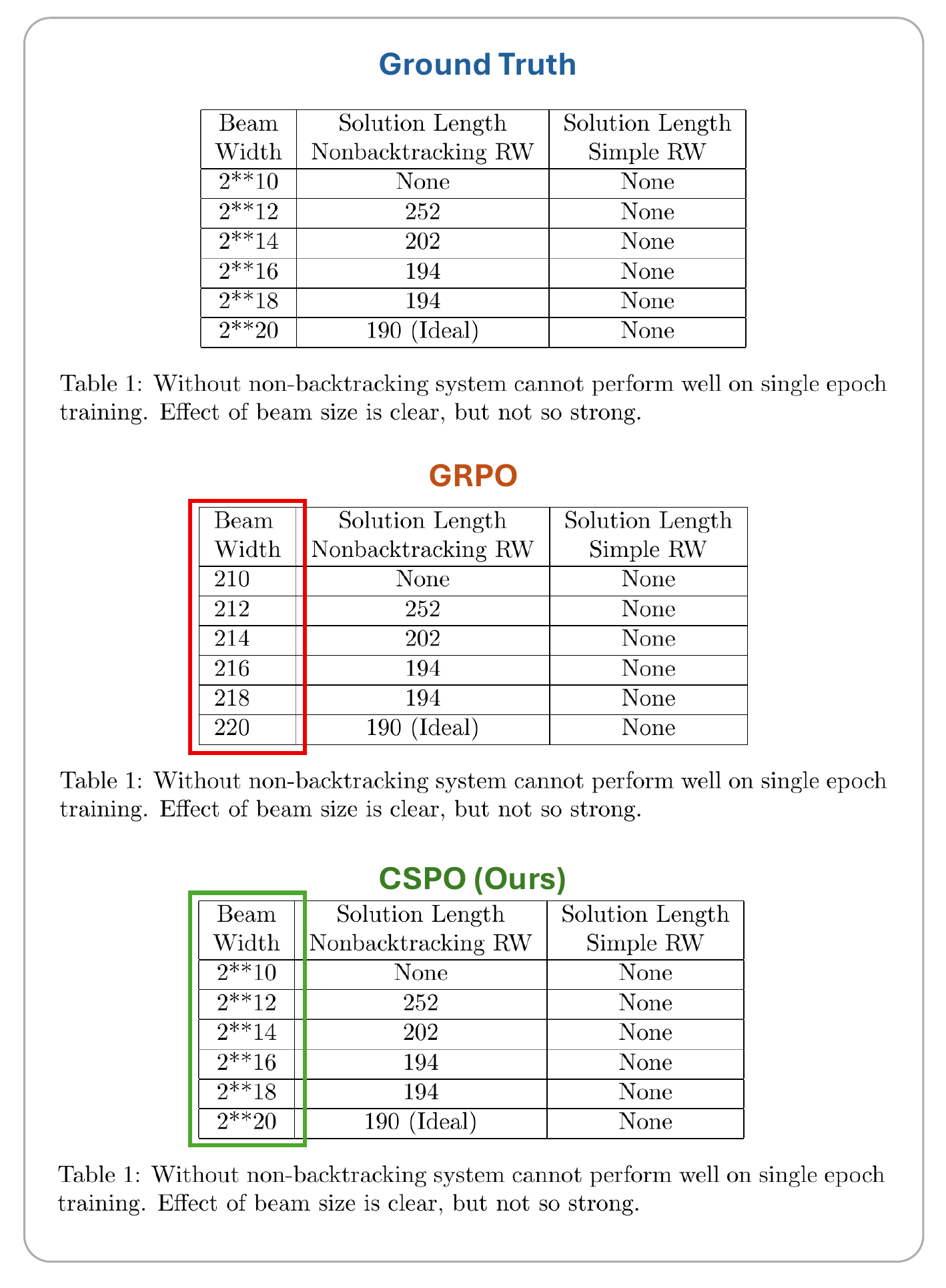}
    \caption{A typical example comparing GRPO and CSPO of 3B models, showing CSPO mitigates \textbf{content errors}.}
    \label{fig:sample-content}
\end{figure}

\begin{figure}[t]
    \centering
    \includegraphics[width=0.95\linewidth]{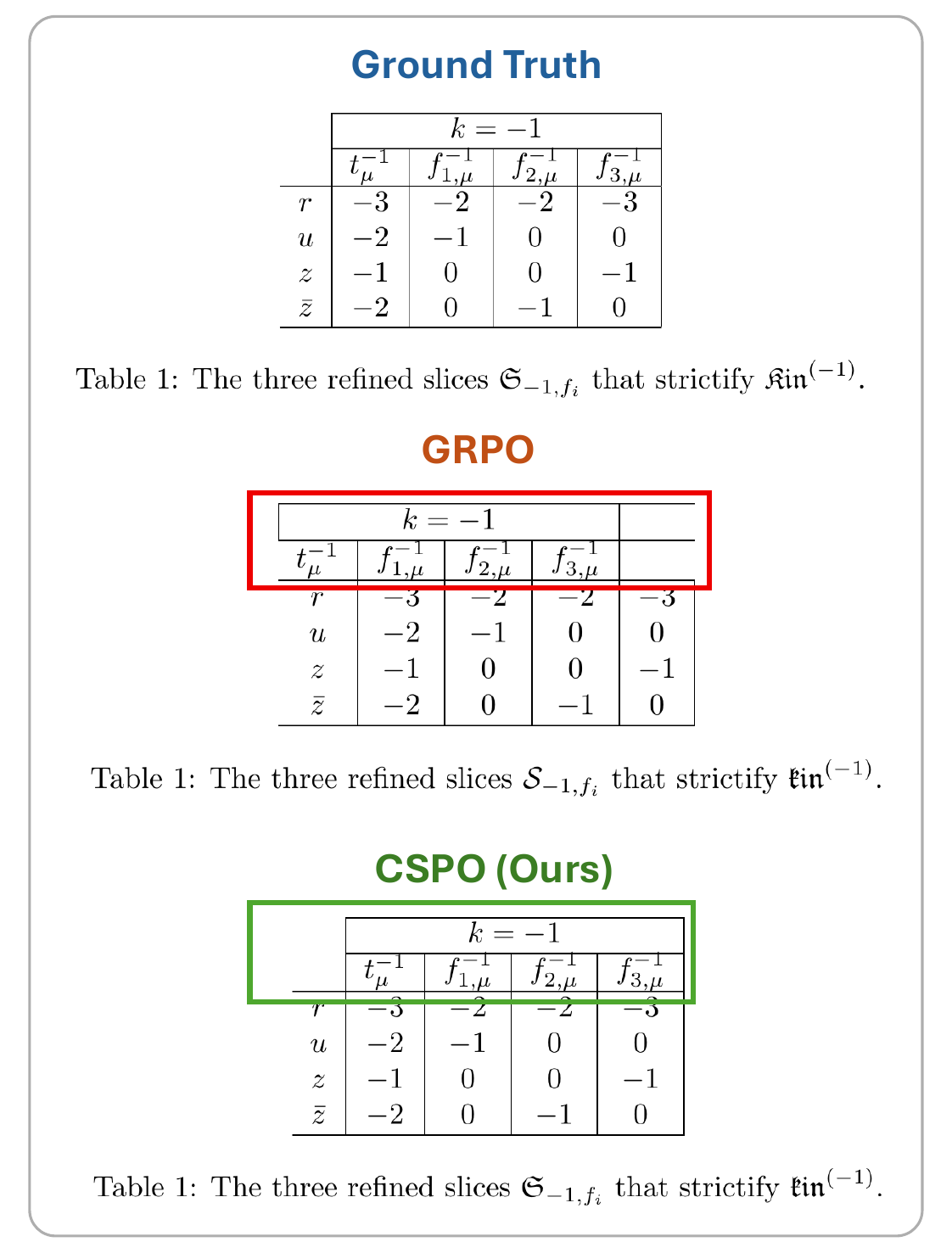}
    \caption{A typical example comparing GRPO and CSPO of 3B models, showing CSPO mitigates \textbf{structure}, and \textbf{content} (in table caption) errors.}
    \label{fig:sample-structure}
\end{figure}

\begin{figure}[t]
    \centering
    \includegraphics[width=0.95\linewidth]{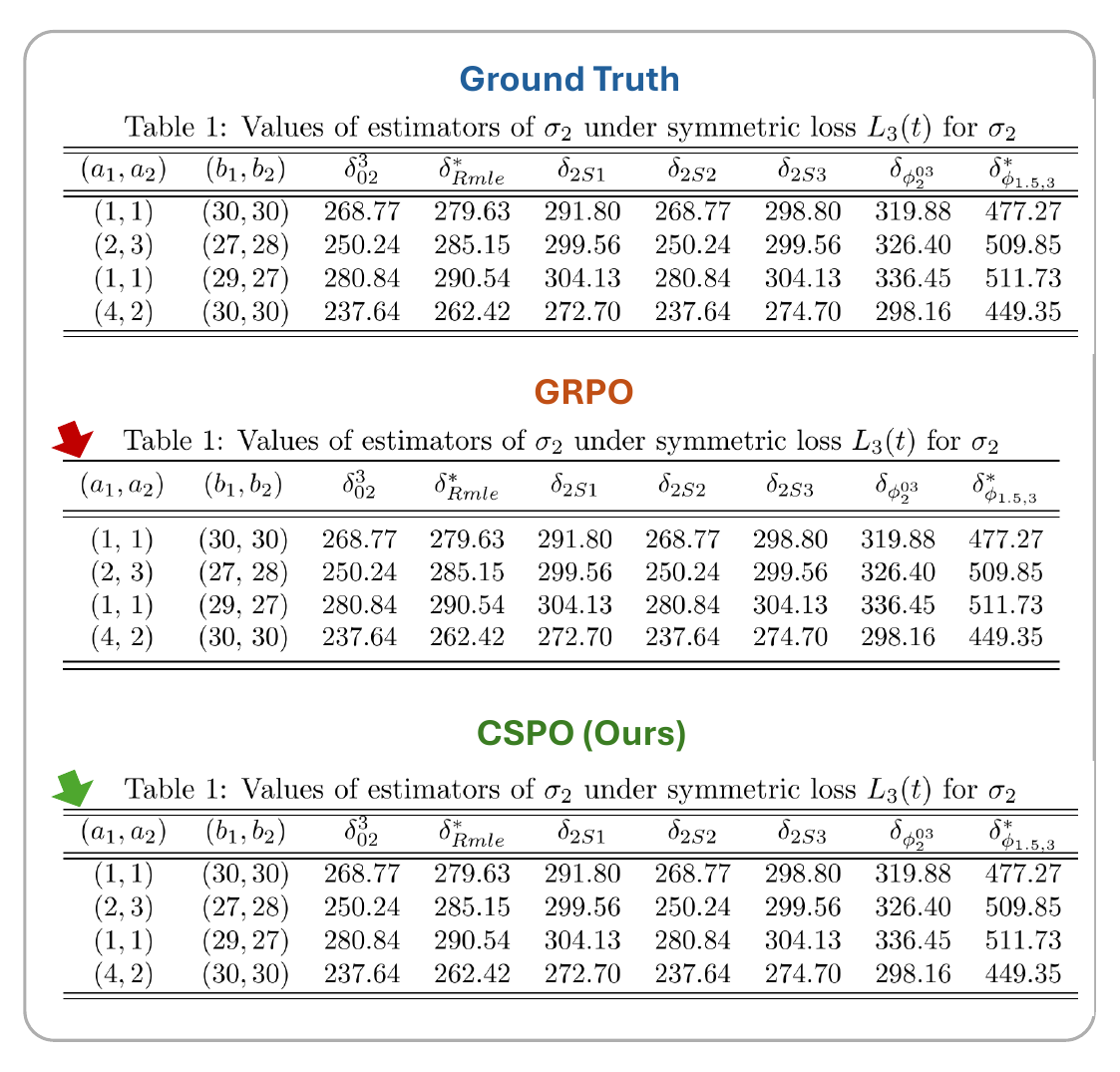}
    \caption{A typical example comparing GRPO and CSPO of 3B models, showing CSPO mitigates \textbf{line style errors}.}
    \label{fig:sample-linestyle}
\end{figure}

\begin{figure}[t]
    \centering
    \includegraphics[width=0.95\linewidth]{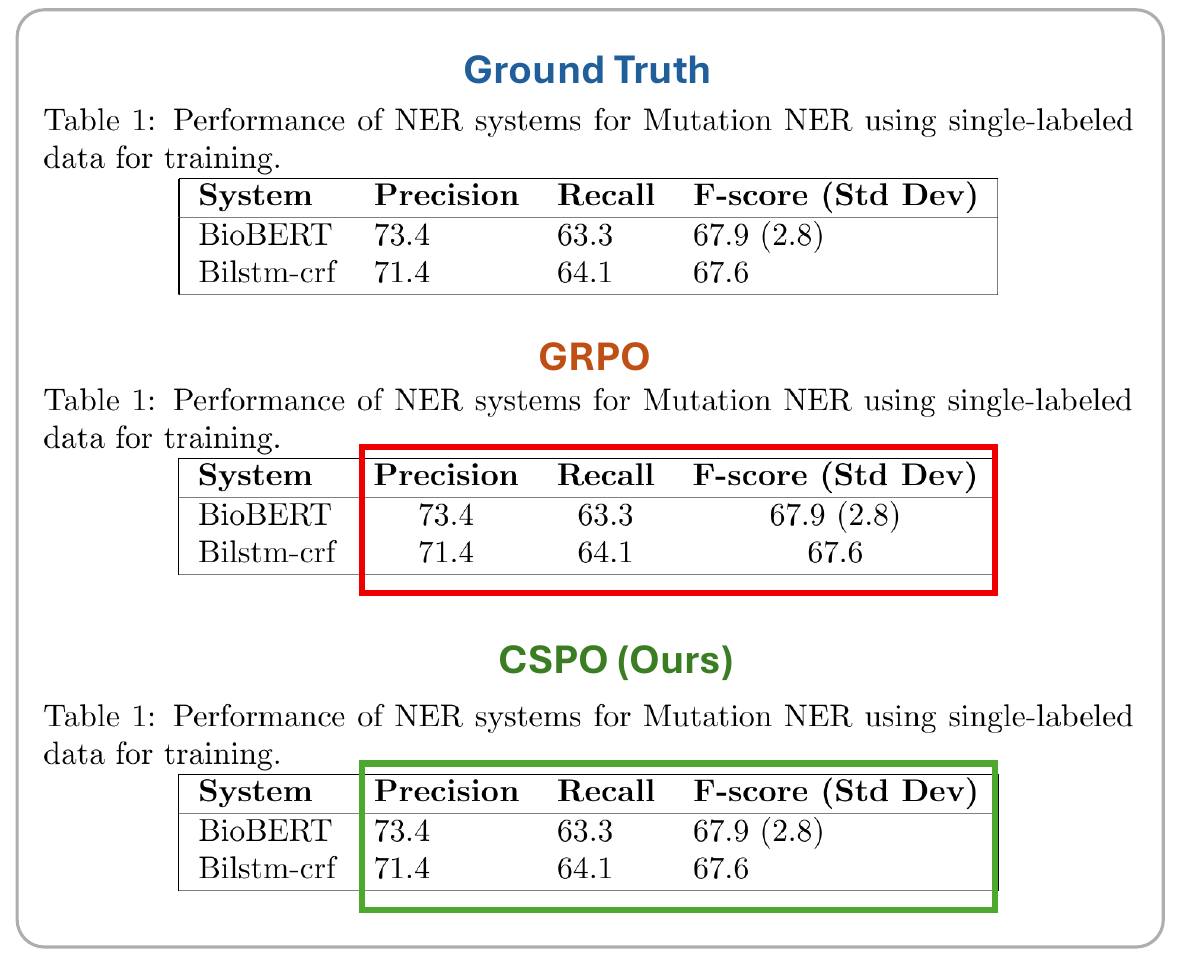}
    \caption{A typical example comparing GRPO and CSPO of 3B models, showing CSPO mitigates \textbf{alignment errors}.}
    \label{fig:sample-alignment}
\end{figure}

\begin{figure}[t]
    \centering
    \includegraphics[width=0.95\linewidth]{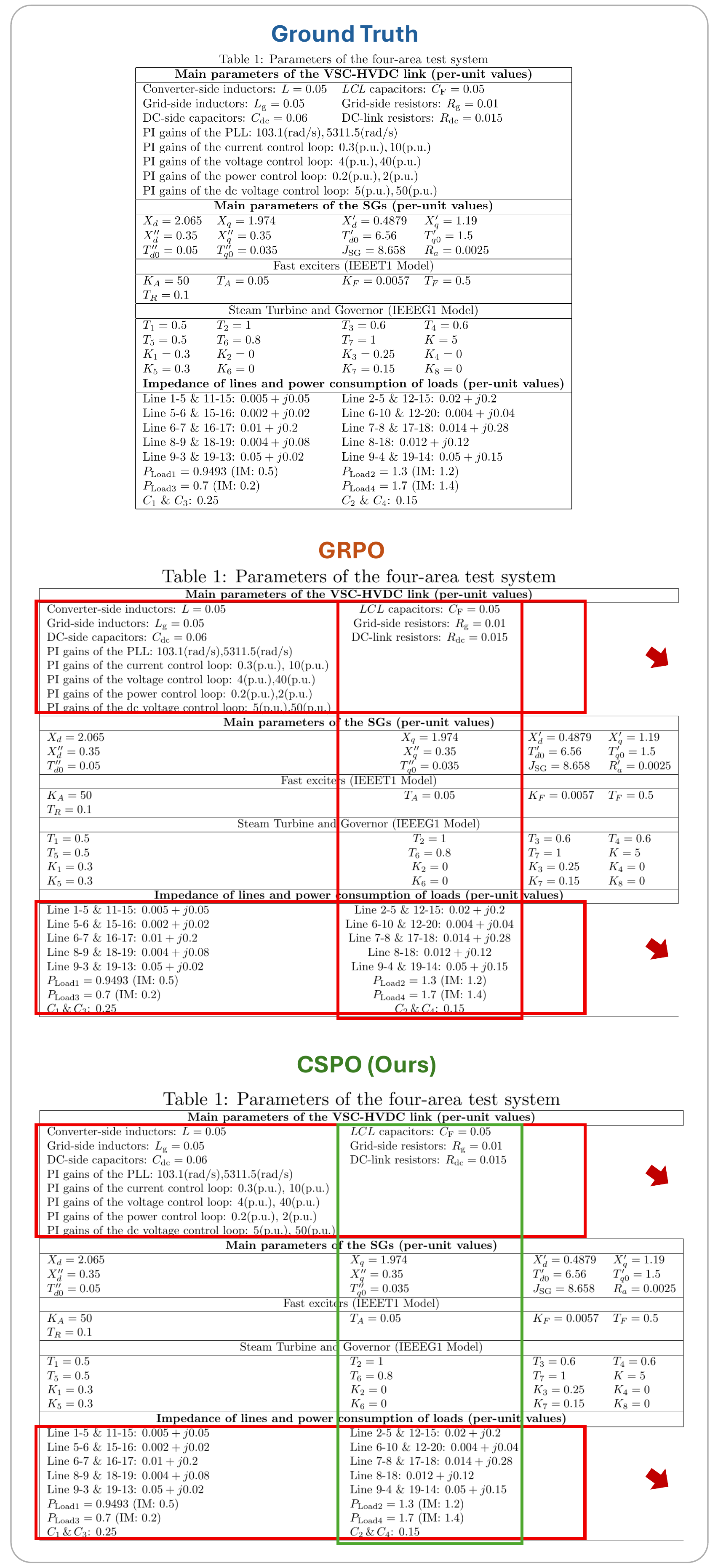}
    \caption{Failure case for GRPO and CSPO of 3B models. Both models' generations exhibit \textbf{structure errors} (marked by red boxes) on this complex table, where \texttt{\textbackslash multicolumn\{\}} is used in groudtruth code but is ignored in the generated code.  In addition, GRPO generation further has \textbf{alignment errors} (center aligned rather than left aligned as groudtruth).}
    \label{fig:failure-case}
\end{figure}

Figure~\ref{fig:sample-content} to Figure~\ref{fig:sample-alignment} show more visualization that CSPO mitigates content, structure, and style errors. Figure~\ref{fig:failure-case} presents a failure case of CSPO on a complex table.

\section{More Discussion}

\subsection{Generalization to Other Tasks}
We validate the proposed component-specific policy optimization framework on the table image-to-LaTeX generation task. The approach is conceptually generalizable to other structured generation problems (\egno, HTML/CSS, code, and presentation generation). 

Our framework includes a \emph{domain-agnostic} credit assignment mechanism and a \emph{domain-aware} component parser. The parser decomposes structured outputs into functional components, enabling localized reward attribution and optimization. Such decomposition is naturally supported in many domains through existing tooling—for example, DOM trees for HTML/CSS, abstract syntax trees (ASTs) for programming languages, and XML-based structures for document formats.

% The credit assignment framework is domain-agnostic while the component-parser is domain-aware. The parser simply maps structured outputs into functional components. For other domains, similar tooling exists (e.g., DOM for HTML/CSS, AST for Python charts, XML parsers for presentations). 

% credit assignment framework is conceptually domain-agnostic, its practical deployment relies on a domain-specific parser (e.g., for LaTeX, HTML, or Python AST), which requires additional engineering effort when extending to other structured generation tasks.

\subsection{Training Cost}

Training CSPO (2 epochs) requires 100+ million tokens in total. We view this as a one-time alignment cost. Actually, our framework is not tied to GPT-4o. It supports local open-weight LLM judges, eliminating API costs. For example, Qwen3-Next-80B achieves ~90\% agreement with human experts in evaluation and demonstrates evaluation trends highly consistent with GPT-4o, providing a scalable and cost-effective alternative for deployment.

% In our setup, training requires on the order of hundreds of millions of tokens, corresponding to a cost on the order of several hundred USD under standard API pricing. While the framework can be implemented with open-weight LLM judges (e.g., Qwen3-Next-80B), which provide comparable evaluation trends and strong agreement with human annotations, the additional cost of LLM-based evaluation remains a practical consideration. (iv)
 
% \section{Data Consent}
% We collect data from arXiv, by using only papers with CC BY, CC BY-SA, CC0, and CC BY-NC licenses. We will  

%\clearpage

\end{document}